\documentclass[smallextended]{svjour3} 
\smartqed 
\journalname{myjournal}
\usepackage{graphicx}
\usepackage{bookmark}
\usepackage{natbib}
\usepackage{amssymb}
\usepackage{amsmath}
\usepackage[ruled,vlined,linesnumbered]{algorithm2e}
\usepackage{colortbl}
\usepackage{chemfig}
\usepackage{subcaption}
\usepackage{bm}
\usepackage{todonotes}
\usepackage{placeins}
\usepackage{centernot}

\newcommand{\Xm}{\mathbf X}  
\newcommand{\Ym}{\mathbf Y}

\begin{document}

\titlerunning{Grouped Feature Importance and Combined Features Effect Plot}

\author{Quay Au\textsuperscript{*}        \and
        Julia Herbinger\textsuperscript{*}  \and
        Clemens Stachl \and
        Bernd Bischl \and
        Giuseppe Casalicchio
}

\authorrunning{Au and Herbinger}
\institute{Q. Au$^1$ \and J. Herbinger$^2$ \and B. Bischl$^3$ \and G. Casalicchio$^4$ \at Department of Statistics,
                Ludwig-Maximilians-University Munich,
                 80539 Munich, Germany \\
              \email{quayau@gmail.com, julia.herbinger@stat.uni-muenchen.de, bernd.bischl@stat.uni-muenchen.de, giuseppe.casalicchio@stat.uni-muenchen.de}           
           \and
           C. Stachl$^5$ \at
        Department of Communication,
       Stanford University,
       94305 Stanford, CA, USA\\
       \email{cstachl@stanford.edu} 
       \and \textbf{*} These authors contributed equally to this work.\\
       \and {CRediT taxonomy: 
       Conceptualization: 1, 2, 3, 4; Methodology: 1, 2, 4; Formal analysis and investigation: 1, 2, 4; Writing - original draft preparation: 1, 2; Writing - review and editing: 3, 4, 5; Investigation: 1, 2; Visualization: 1, 2; Validation: 1, 2, 4; Software: 1, 2; Funding acquisition: 3, 5; Supervision: 3, 4}
}

\title{Grouped Feature Importance and Combined Features Effect Plot}

\date{Received: date / Accepted: date} 

\maketitle

\begin{abstract}
Interpretable machine learning has become a very active area of research due to the rising popularity of machine learning algorithms and their inherently challenging interpretability. Most work in this area has been focused on the interpretation of single features in a model. However, for researchers and practitioners, it is often equally important to quantify the importance or visualize the effect of feature groups. To address this research gap, we provide a comprehensive overview of how existing model-agnostic techniques can be defined for feature groups to assess the grouped feature importance, focusing on permutation-based, refitting, and Shapley-based methods. We also introduce an importance-based sequential procedure that identifies a stable and well-performing combination of features in the grouped feature space. Furthermore, we introduce the combined features effect plot, which is a technique to visualize the effect of a group of features based on a sparse, interpretable linear combination of features. We used simulation studies and a real data example from computational psychology to analyze, compare, and discuss these methods.  
\keywords{Grouped Feature Importance \and Combined Features Effects \and Dimension Reduction \and Interpretable Machine Learning}
\end{abstract}


\newpage

\section{Introduction}

The popularity of machine learning (ML) algorithms has grown considerably in recent years, especially because they have often demonstrated outstanding performance in modeling complex and non-linear relationships. 
ML algorithms are nowadays used in many diverse fields such as medicine \citep{shipp2002diffuse}, criminology \citep{berk2009forecasting}, 
and increasingly in the social sciences \citep{Stachl2020ML, Yarkoni2017}.
Well-performing ML models often come along with a lack of interpretability.
However, interpretable models are paramount in many high-stakes settings such as medical and juridical applications \citep{lipton2018mythos}.
In the context of interpretable ML (IML) research, several model-agnostic methods to understand the influence of a single feature's importance or effect have been developed \citep{molnar2019IML}.
Examples include the permutation feature importance \citep[PFI;][]{modelreliance2018}, leave-one-covariate out (LOCO) importance \citep{Jing2018LOCO}, SHAP values \citep{lundberg:2017}, or partial dependence plots \citep[PDP;][]{friedman2001greedy}. 

In many applications, it can be more informative to quantify the importance or effect of a group of features. 
From a computational and run-time perspective, this might be more efficient and relevant for high-dimensional datasets, especially when groups of features can either be defined in a data-driven or in a knowledge-driven way \citep{He2010}. In data-driven grouping, an algorithmic approach can be used to define groups of features, which can be useful in the case of highly correlated feature spaces such as in genetic applications \citep{Park2006, Tolosi2011}. 
Besides the computational advantage, 
there are theoretical and practical reasons in favor of the grouped feature perspective. 
One such reason is that many IML techniques rely on the assumption of independent features.
Hence, applying these methods to individual features might lead to misleading results. 
However, features can be grouped in such a way that the underlying assumptions hold, yielding more meaningful interpretations.
If groups can be naturally defined by the user (knowledge-driven grouping) such as 
in applications with sensor data \citep{Chakraborty2008}, quantifying or visualizing the influence of feature groups might be more informative or might lead to additional insights. There are also use cases where the interpretation of single features might be misleading. Examples include datasets with time-lagged or categorical features (e.g., one-hot encoded categories) and the presence of feature interactions \citep{Gregorutti2015}. 

Although the grouped feature perspective is relevant in many applications, most of the IML research has focused on methods that try to provide explanations on a single feature level. Model-agnostic methods for feature groups are rare and not well-studied. 
We provide a comprehensive overview and extensions of available approaches and introduce two new methods. A sequential grouped feature importance procedure and the combined features effect plot (CFEP) which visualizes the effect of a group of features on the prediction.

\subsection{Related Work}
 

A well-known model that handles grouped features is the group LASSO \citep{Yuan2006},
which extends the LASSO \citep{tibshirani1996regression} for feature selection based on groups.
Moreover, other extensions, e.g., to obtain sparse groups of features \citep{friedman2010note}, to support classification tasks \citep{meier2008group} or non-linear effects \citep{gregorova:2018} exist.
However, group LASSO is a modeling technique that focuses on selecting groups in the feature space rather than quantifying their importance. 

A large body of research already exists regarding the importance of individual features \citep[see, e.g.,][]{modelreliance2018, hooker2019}. 
\cite{hooker2019} distinguish between two major approaches to measure the feature importance based on loss functions, namely permutation methods and refitting methods. Permutation methods measure the increase in error after permuting a feature while the model remains untouched. Refitting methods measure the increase in error after leaving out the feature of interest completely and refitting the model.
\cite{Gregorutti2015} introduced a model-specific, grouped permutation feature importance (PFI) score for random forests and applied this approach to functional data analysis.
\cite{valentin2020interpreting} introduced a model-agnostic grouped version of the model reliance score \citep{modelreliance2018}.
However, they are focusing more on the application and leaving out a detailed theoretical foundation. 
Recently, a general refitting framework to measure the importance of (groups of) features was introduced by \cite{williamson2020unified}. 
In their approach, the feature importance measurement is detached from the model level and defined by an algorithm-agnostic version to measure the intrinsic importance of features. 
The importance score is defined by the difference between the performance of the full model and the performance based on all features except the group of interest.

While permutation methods have the advantage that evaluations are often cheaper than those of refitting methods, it has been shown that PFI often fails when features are dependent since the method extrapolates in regions without any or just a few observations \citep{hooker2019}. 
Hence, interpretations in these regions might be misleading. 
To avoid this problem, alternatives based on conditional distributions or refitting have been suggested \citep[e.g.,][]{Strobl2008, nicodemus2010, hooker2019, Watson2019, molnar2020}. 
Although the conditional PFI provides a solution to this problem, the interpretation of the score changes.
Conditional PFI can only be interpreted depending on the underlying conditional distribution, meaning it ``must be interpreted as the additional, unique contribution of a feature given all remaining features we condition on were known'' \citep{molnar2020}. 
This property complicates the comparison with non-conditional interpretation methods. Therefore, we do not consider any conditional variants in this paper.
Refitting methods, on the other hand, are computationally more expensive, in particular in a large feature space.

A third group of importance measures is based on Shapley values \citep{shapley1953value}, a theoretical concept of game theory. The additive feature importance measure SHAP \citep{lundberg:2017} quantifies the attribution of each feature to the predicted outcome and refers to a permutation-based method. It has the advantage
that contributions of interactions are distributed fairly between features. Besides being computationally more expensive, SHAP itself is based on the model's outcome rather than its performance. \citet{casa2019} extended the concept of Shapley values to fairly distribute the model's performance among features and called it Shapley Feature IMPortance (SFIMP). A similar approach has also been proposed by \citet{covert2020SAGE}, who 
showed the benefits of the method on various simulation studies. 
One approach that uses Shapley values to explain grouped features was introduced by \citet{mijolla:2020}. However, they derived latent variables of the feature groups and applied an adjusted version of Shapley values on these variables instead of using the underlying features themselves. Also, \citet{amoukou2021shapley} investigated grouping approaches for Shapley values in the case of encoded categorical features and subset selection of important features for tree-based methods.
The calculation of Shapley values on groups of features based on performance values has only been applied with regards to feature subset selection methods and not for interpretation purposes \citep{Shay:2005, tripathi2020interpretable}.


After identifying which groups of features are important, the user is often interested in how they (especially the important groups) influence the model's prediction. Therefore, several techniques to visualize single-feature effects exist. 
These include partial dependence plots (PDP) \citep{friedman2001greedy}, individual conditional expectation (ICE) curves \citep{icecurves}, SHAP dependence plots \citep{lundberg:2018}, and accumulated local effects (ALE) plots \citep{apley2016visualizing}. While the first three mentioned methods inherit similar disadvantages as PFI, ALE is more suitable if features are correlated. However, in the case of high-dimensional feature spaces, it is not feasible for the user to compute, visualize and interpret single-feature plots for all (important) features.
If features are grouped, visualization techniques become computationally more complex and it may become even harder to visualize the results in an easily interpretable way. In the case of low-dimensional feature spaces, this might still be feasible, for example by using two-features PDPs or ALE plots.
Recently, effect plots that  visualize the combined effect of multiple features have been introduced by \citet{seedorff2021totalvis} and \citet{brenning2021transforming}. They use PCA to reduce the dimension of the feature space and calculate marginal effect curves for the principal components. However, the used dimension reduction method lacks in including information about the target variable and lacks in sparsity and hence interpretability. 

\subsection{Contribution}



Our contributions can be summarized as follows: We extend the permutation-based and refitting grouped feature importance methods introduced by \cite{valentin2020interpreting} and \cite{williamson2020unified} by not only comparing to the full model (i.e., taking into account all features) but also to a null model (i.e., ignoring all features). Hence, we can quantify how much a group itself contributes to the prediction of a model without the presence of other groups. Furthermore, we introduce Shapley importance for feature groups and how these scores can be decomposed into single-feature importance scores of the respective groups. Moreover, we define a new algorithm to sequentially add groups of features depending on their importance and with that being able to find well-performing combinations of groups. We compare all methods regarding the main challenges that arise when quantifying grouped feature importance by creating small simulation examples. Therefore, we provide recommendations for using and interpreting the respective methods correctly. 
The main challenges are finding good and sparse combinations of features (i.e., groups) when dependencies between groups are present, managing varying correlations within groups of features, and handling varying group sizes. 
Finally, we introduce a model-agnostic method to visualize the joint effect of a group of features. Hereby, we use a suitable dimension reduction technique and the conceptual idea of PDPs to calculate and plot the mean prediction of a sparse group of features regarding their linear combination. This novel method finally enables the user to visualize effects for groups of features.
We showcase the usefulness of all these methods in a real data example from computational psychology.

The structure of this paper is as follows: In Section \ref{sec:featimp}, we formally define the grouped feature importance methods and introduce the sequential grouped feature importance procedure. We compare these methods for different scenarios in Section \ref{sec:challenges}. In Section \ref{sec:groupedEffect}, we introduce the CFEP to visualize the effects of feature groups based on a supervised dimension reduction technique. Therefore, we also show the suitability of this technique compared to its unsupervised counterpart in a simulation study. Finally, in Section \ref{sec:realdata}, all methods are applied to a real data example before summarizing and giving a prospect in Section \ref{sec:conclusion}.

\subsection{Reproducibility and Open Science}
The implementation of the proposed methods and reproducible scripts for the experimental analysis are provided in the following public git-repository \url{https://github.com/JuliaHerbinger/grouped_feat_imp_and_effects}.

\section{Feature Importance for Groups}
\label{sec:featimp}
Analogous to \cite{casa2019}, we use the term feature importance as the influence of a feature or a group of features on a model's predictive performance, which we measure by the expected loss when we perturb these features in a permutation approach or remove these features in a refitting approach.

In the upcoming chapters, we provide a general notion and formal definitions for permutation and refitting methods and explain them by answering the following questions:
\begin{itemize}
 \setlength{\itemsep}{0pt}\setlength{\parskip}{0pt}
    \item[a)] How much does a group of features contribute to the model's performance in the presence of other groups? 
    \item[b)] How much does a group itself increase the expected loss if it is added to a null model like the mean prediction of the target for refitting methods?
    \item[c)] How can we fairly distribute the expected loss among all groups and all features within a group? 
    \item[d)] How can we find well-performing combinations of groups?
\end{itemize}
The definitions of all grouped feature importance scores are based on loss functions. They are defined in such a way that important groups will yield positive grouped feature importance scores.
The question of how to interpret the differing results of these methods is addressed in Section \ref{sec:challenges}.


\subsection{General Notation}
\label{Notation}
Consider a $p$-dimensional feature space $\mathcal X = (\mathcal X_1 \times ... \times \mathcal X_p)$ and a one dimensional target space $\mathcal Y$. The corresponding random variables, which are generated from these spaces are denoted by $ X = ( X_1, ...,  X_p)$ and $ Y$.
Furthermore, assume that there is an unknown functional relation $f : \mathcal X \longrightarrow \mathcal Y$.
ML algorithms try to learn this functional relationship using $n \in \mathbb N$ i.i.d. observations drawn from the joint space $\mathcal X \times \mathcal Y$ with unknown probability distribution $\mathcal{P}$. We denote this by the dataset  $\mathcal D = \{(\mathbf x^{(i)}, y^{(i)})\}_{i = 1}^n$, where the vector $\mathbf x^{(i)} = (x_1^{(i)}, ..., x_p^{(i)})^\intercal \in \mathcal X$ is the $i$-th observation associated with the target variable $y^{(i)} \in \mathcal Y$.
The $j$-th feature is denoted by $\mathbf x_j = (x_j^{(1)}, ..., x_j^{(n)})^\intercal$, for $j = 1, ..., p$.
The dataset $\mathcal D$ can also be written in matrix form
\begin{equation}
\label{datamatrix}
\begin{pmatrix}
x_{1}^{(1)} & \hdots &  x_{p}^{(1)}& y^{(1)}\\ 
\vdots & \ddots & \vdots & \vdots\\ 
x_{1}^{(n)} & \hdots & x_{p}^{(n)} & y^{(n)}
\end{pmatrix}
  = 
\begin{pmatrix}
\Xm, \Ym  
\end{pmatrix}
\text{, with }
\Xm = \begin{pmatrix}
x_{1}^{(1)} & \hdots &  x_{p}^{(1)}\\ 
\vdots & \ddots & \vdots\\ 
x_{1}^{(n)} & \hdots & x_{p}^{(n)}
\end{pmatrix}
\text{ and }
\Ym = \begin{pmatrix}
 y^{(1)}\\ 
\vdots\\ 
y^{(n)}
\end{pmatrix}.
\end{equation}

The generalization error $GE(\hat f, \mathcal P) = \mathbb E(L(\hat f (X), Y))$ of a learned model $\hat f$ is measured by a loss function $L$ on test data drawn independently from $\mathcal P$.
Hence, in this case, the generalization error is defined by the expected loss of a learned model and therefore can be estimated by taking the mean of the loss function on unseen test data $\mathcal D_{\text{test}}$
\begin{equation}
    \label{eq:ge2}
    \widehat{GE}(\hat f, \mathcal D_{\text{test}}) = \frac{1}{|\mathcal D_{\text{test}}|}\sum_{(\mathbf x, y) \in \mathcal D_{\text{test}}} L(\hat f(\mathbf x), y).
\end{equation}
The application of an algorithm $a$ to a given dataset $\mathcal D$ results in a fitted model $a(D) = \hat f_{D}$.
The \textit{expected generalization error} of an algorithm $a$ takes into account the variability introduced by sampling different datasets $\mathcal D$ of equal size $n$ from $\mathcal{P}$ and is defined by
\begin{equation}
\label{eq:ege}
GE(a, \mathcal{P}, n) = \mathbb E_{|\mathcal D| = n}(GE(a(\mathcal D), \mathcal{P})).
\end{equation}
In practice, resampling techniques on the available dataset $\mathcal D$ are used to estimate Eq.~\eqref{eq:ege}.
Resampling techniques usually split the dataset $\mathcal D$ into $k \in \mathbb N$ training datasets $\mathcal D_{\text{train}}^i$, $i = 1,...,k$, of roughly the same size $n_{\text{train}} < n$.  The estimate of an algorithm's generalization error is the average of the estimations on each test dataset
\begin{equation}
\label{eq:ege2}
\widehat{GE}(a, \mathcal D, n_{\text{train}}) = \frac{1}{k} \sum_{i=1}^{k} \widehat{GE}(\hat f_{\mathcal D_{\text{train}}^i}, \mathcal D_{\text{test}}^i).
\end{equation}
In the following, we often associate the set of numbers $\{1, ..., p\}$ in a one-to-one manner with the features $\mathbf x_1, ..., \mathbf x_p$ and refer the number $i \in \{1, ...,p\}$ as feature $x_i$.
We call $G \subset \{1, ..., p\}$ a \textit{group of features}.






\subsection{Permutation Methods}
Inspired by the PFI measure used in random forests \citep{Breiman2001}, \cite{modelreliance2018} proposed a model-agnostic version. 
The PFI score of feature $j$ of a fitted model $\hat f$ is defined as the increase in expected loss after permuting the feature values
\begin{equation}
\label{eq:PFI}
\text{PFI}_j(\hat f) = \mathbb E(L(\hat f (X_{[j]}), Y)) - \mathbb E(L(\hat f(X), Y)).
\end{equation}
Here, $X_{[j]} = (X_1, ..., X_{j-1}, \tilde X_j, X_{j + 1}, ..., X_p)$ is the $p$ dimensional random variable vector of features, where $\tilde X_j$ is an independent replication of $X_j$. The random variable $\tilde X_j$  has the same distribution as $X_j$, but is independent of all other features and the target variable. In practice, this is done by permuting the values in the data column of the $j-$th feature.
The idea behind this method is to break the association between the $j-$th feature and the target variable by permuting the feature values. 
If a feature is not useful for predicting an outcome, changing its values by permuting, will not increase the expected loss. The larger the PFI score of feature $j$, the more substantial the increase in error and the more important the considered feature\footnote{Here, we solely consider the case of loss functions that are to be minimized.}.

This procedure can be performed for each feature $j = 1, ..., p$ to quantify the respective importance scores. 
The estimation relies on the repeated permutation of a feature for a predefined number of times to form a distribution of importance scores. 
On feature level, these scores are summarized by their means to provide a final score that can be compared across different features.
For an accurate estimation of Eq.~\eqref{eq:PFI}, we would need to calculate all possible permutation vectors over the observation index set $\{1, ..., n\}$, see also \cite{casa2019} for an in-depth discussion on this topic. However, Eq.~\eqref{eq:PFI} can also be approximated on a dataset $\mathcal D$ with $n$ observations by Monte Carlo integration using $m$ random permutations:
\begin{equation}
\label{eq:hatpfi}
    {\scriptstyle \widehat{\text{PFI}}_j(\hat{f}, \mathcal D) = \frac{1}{n m} \sum_{i = 1}^n \sum_{k = 1}^m \left(L\left(\hat f(\mathbf (x^{(i)}_1, ..., x^{(\tau_k^{(i)})}_j, ... ,x^{(i)}_{p}), y^{(i)})\right) - L\left(\hat f(\mathbf x^{(i)}, y^{(i)})\right) \right),}
\end{equation}
where $\tau_k$ is a random permutation vector of the index set $\{1, ..., n\}$ for $k = 1,...,m$. An example for $n = 3$ would be $\tau_1 = (1, 3, 2)^\intercal$ with $\tau_1^{(i)}$ being the $i-$th entry of that vector.

It should be noted that the PFI measure in random forest models is computed on naturally occurring out-of-bag samples \citep{Breiman2001}.
The procedure above could also be embedded into a resampling technique, where the permutation is always applied on the held-out test set of each resampling iteration \citep{modelreliance2018}. However, this leads to refits and is computationally more expensive.
The resulting resampling-based PFI is estimated by
\begin{equation}
\label{eq:PFIresampling}
\widehat{\text{PFI}}_j^{\text{res}}(a, \mathcal D, n_{\text{train}}) = \frac{1}{k}\sum_{i = 1}^k \widehat{\text{PFI}}_j(\hat f_{\mathcal D_{\text{train}}^i}, \mathcal D_{\text{test}}^i),
\end{equation}
where the permutation strategy is applied on the test sets $\mathcal D_{\text{test}}^i$. 
In the following, we extend this existing definition of permutation importance to groups of features and introduce the GPFI (Grouped Permutation Feature Importance) and GOPFI (Group Only Permutation Feature Importance) scores. For ease of notation, we will only define these scores for a given model (see Eq. (\ref{eq:PFI})).

\subsubsection{Grouped Permutation Feature Importance}
For the definition of GPFI which is based on the definitions of \cite{Gregorutti2015} and \cite{valentin2020interpreting}, let $G \subset \{1, ..., p\}$ be a group of features.
With slight abuse of notation to index the feature groups included in $G$, we define the grouped permutation feature importance of $G$ as
\begin{equation}
\label{GPFI}
\text{GPFI}_{G} = \mathbb E(L(\hat f (\tilde X_{G}, X_{-G}), Y)) - \mathbb E(L(\hat f(X), Y)).
\end{equation}
Here, $\tilde X_{G} = (\tilde X_j)_{j \in G}$ is a $|G|$-dimensional random vector of features, which is an independent replication of $X_{G} = (X_j)_{j \in G}$. Also this random vector is independent of both the target variable and the random vector of remaining features, which we define by $X_{-G} := (X_j)_{j \in \{1, ..., p\} \backslash G}$.
It extends Eq.~\eqref{eq:PFI} to groups of features so that the interpretation of GPFI scores always refers to the importance when the feature values of the group defined by $G$ are permuted jointly (i.e., without destroying the dependencies of the features within the group).
Similar to Eq.~\eqref{eq:PFIresampling}, the grouped permutation feature importance can be estimated by monte carlo integration:
\begin{equation}
\label{gpfi}
\widehat{\text{GPFI}}_{G} = \frac{1}{nm} \sum_{i = 1}^n \sum_{k = 1}^m\left(L(\hat f(\mathbf x_{G}^{(\tau_k^{(i)})}, \mathbf x_{-G}^{(i)}), y^{(i)}) - L(\hat f(\mathbf x^{(i)}, y^{(i)})) \right).
\end{equation}
The GPFI measures the contribution of one group to the model's performance if all other groups are present in the model (see (a) from Section \ref{sec:featimp}). 

\subsubsection{Group Only Permutation Feature Importance}
To evaluate how much a group itself contributes to a model's performance, one can also use a slightly different measure.
As an alternative to Eq.~\ref{gpfi}, we can also compare the expected loss after permuting all features jointly with the expected loss after permuting all features except the considered group.
We define this group only permutation feature importance (GOPFI) for a group $G \subset \{1, ..., p\}$ as
\begin{equation}
\label{GOPFI}
\text{GOPFI}_{G} =  \mathbb E(L(\hat f(\tilde X), Y)) - \mathbb E(L(\hat f (X_{G}, \tilde X_{-G}), Y)),
\end{equation}
which can be approximated by
\begin{equation}
\label{gopfiapprox}
\widehat{\text{GOPFI}}_{G} = \frac{1}{nm} \sum_{j = 1}^n \sum_{k = 1}^m\left(L(\hat f(\mathbf x^{(\tau_k^{(j)})}, y^{(j)})) - L(\hat f(\mathbf x_{G}^{(j)}, \mathbf x_{-G}^{(\tau_k^{(j)})}), y^{(j)}) \right).
\end{equation}
Furthermore, GOPFI is technically useful for the permutation variant of the Shapley importance (see Eq. (\ref{eq:shapvaluefunction})).

\subsection{Refitting Methods}
Another possibility to determine the importance of features is based on refitting. Permutation methods do not require any refits to calculate the importance scores. Hence, they are often computationally cheaper to compute than refitting methods. However, since the model remains untouched in the former approach, interpretations are solely based on the specific model, while interpretations for refitting methods can be generalized to the underlying algorithm. In \cite{Jing2018LOCO} a model-agnostic feature importance measure, namely leave-one-covariate-out (LOCO), was introduced. This approach calculates the feature importance of single features by removing them and refitting the model. The feature importance value is defined as the difference in expected loss between the full model and the model that was fitted on the reduced dataset.
In situations with many features, this can quickly become computationally challenging, since for each feature and every resampling iteration a separate model has to be fit. However, if the features are being grouped and the number of groups is reasonably small, this method can be feasible in many applications. 
In the following chapters, we will introduce two LOCO-based refitting methods for groups of features. The first definition is similar to the one introduced in \cite{williamson2020unified}.

\subsubsection{Leave-One-Group-Out Importance}
For a subset $G \subset \{1, ..., p\}$, we define the reduced dataset $\tilde{\mathcal{D}} :=  \{(\mathbf x_{-G}^{(i)}, y^{(i)})\}_{i = 1}^n$. Given an algorithm $a$, which generates models $a(\mathcal{D}) = \hat f_{\mathcal{D}}$ and $a(\tilde{\mathcal{D}}) =  \hat f_{\tilde{\mathcal{D}}}$, we define the Leave-One-Group-Out Importance (LOGO)  as
\begin{equation}
    \label{eq::dgi}
    LOGO(G) = \mathbb E(  L( \hat f_{\tilde{\mathcal{D}}} (X_{-G}), Y ) ) - \mathbb E(  L( \hat f_{\mathcal{D}} (X), Y ) ).
\end{equation}
The LOGO can be estimated by using the algorithm $a$ on $\tilde{\mathcal{D}}$ and should be embedded in a resampling technique:
\begin{eqnarray*}
    \label{eq:dgiestimated}
    \widehat{LOGO}(G) &&= 
\widehat{GE}(a, \tilde{\mathcal{D}}, n_{\text{train}}) - \widehat{GE}(a, \mathcal{D}, n_{\text{train}}) \\
&&= \frac{1}{k} \sum_{i=1}^{k} \widehat{GE}(\hat f_{\tilde{\mathcal{D}}_{\text{train}}^i},\tilde{\mathcal{D}}_{\text{test}}^i) - \frac{1}{k} \sum_{i=1}^{k} \widehat{GE}(\hat f_{ \mathcal{D}_{\text{train}}^i}, \mathcal{D}_{\text{test}}^i).
\end{eqnarray*}
It follows that we compare the loss increase relative to the full model's expected loss when leaving out a group of features and performing a refit. 

\subsubsection{Leave-One-Group-In Importance}
While it may be too limiting to estimate the performance of a model based on one feature only, it can be informative to see how much a group of features(e.g., all measurements from a specific medical device) can reduce the expected loss in contrast to a null model. The Leave-One-Group-In (LOGI) method could be particularly helpful in settings where information on additional groups of measures will inflict significant costs (e.g., adding functional imaging data for a diagnosis) and or limited resources are available (e.g., in order to be cost-covering only one group of measures can be acquired). The LOGI method can also be useful for theory development in the natural and social sciences (e.g., which group of behaviors is most predictive by itself).

Let $a_{\text{null}}$ be a null algorithm, which results in a null model $\hat f_{\text{null}}$, that only guesses the mean (or majority class for classification) of the target variable for any dataset. We additionally define an algorithm $a$, which generates a model $a(\mathring{\mathcal{D}}) = \hat f_{\mathring{\mathcal{D}}}$ for a dataset $\mathring{\mathcal{D}} :=  \{(\mathbf x_{G}^{(i)}, y^{(i)})\}_{i = 1}^n$, which only contains features defined by $G \subset \{1, ..., p\}$. We define the $LOGI$ of a group $G$ as
\begin{equation}
    \label{eq::goi}
    LOGI(G) = \mathbb E(  L( \hat f_{\text{null}}, Y) ) - \mathbb E(  L( \hat f_{\mathring D} (X_{G}), Y ) ).
\end{equation}

The LOGI can be estimated by using the algorithm $a$ on $\mathring{\mathcal{D}} = \{(\mathbf x_{G}^{(i)}, \mathbf y^{(i)})\}_{i = 1}^n$ and should be embedded in a resampling technique:
\begin{eqnarray*}
    \label{eq:goiestimated}
    \widehat{LOGI}(G) &&= 
\widehat{GE}(a_{\text{null}}, \mathcal{D}, n_{\text{train}}) - \widehat{GE}(a, \mathring{\mathcal{D}}, n_{\text{train}}) \\
&&= \frac{1}{k} \sum_{i=1}^{k} \widehat{GE}(\hat f_{\text{null}}, \mathcal{D}_{\text{test}}^i) - \frac{1}{k} \sum_{i=1}^{k} \widehat{GE}(\hat f_{ \mathring{\mathcal{D}}_{\text{train}}^i}, \mathring{\mathcal{D}}_{\text{test}}^i).
\end{eqnarray*}


\subsection{Grouped Shapley Importance}
\label{sec:shapleyImp}
The importance measures defined above either exclude (or permute) individual groups of features from the total set of features or consider only the importance of groups omitting (or permuting) all other features.
The grouped importance scores are usually not affected if interactions within the groups are present. However, they can be affected if features from different groups interact since permuting a group of features jointly destroys any interactions with other features outside the considered group.
We, therefore, define the grouped Shapley importance (GSI) based on Shapley values \citep{shapley1953value}. GSI scores account for feature interactions as they measure the average contribution of a given group to all possible combinations of groups and fairly distribute the importance value caused by interaction values among all groups.

Given a set of groups $\mathcal G = \{G_1, ..., G_l\}$, with $G_i \subset \{1, ..., p\}$, for $i = 1,...,l$.
In our grouped feature context, the value function $v:\mathcal P(\mathcal G) \longrightarrow \mathbb R$ assigns a ``payout'' to each possible group or combination of groups included in $\mathcal G$. With slight abuse of notation, we define the value function for a subset $S \subset \mathcal G$ as
$$
v(S) := v\left(\cup_{G_i \in S} G_i\right).
$$
We define the value function for a group $G \in \mathcal G$ calculated by a refitting or a permutation method by
\begin{equation}
\label{eq:shapvaluefunction}
v_{\text{refit}}(G) = LOGI(G) \text{\hspace{5pt} or \hspace{5pt}} v_{\text{perm}}(G) = GOPFI(G),
\end{equation}
respectively. 
The marginal contribution of a group $G \in \mathcal G$, with $S \subset \mathcal G$ is given by
$$
\Delta_G(S) = v(S\cup G) - v(S).
$$
The GSI of the feature group $G$ is then defined as
\begin{equation}
    \label{eq:shapleygroupformula}
    \phi(G) = \sum_{S \subset \mathcal G \backslash G}\frac{(|\mathcal G| - 1 - |S|)! \cdot |S|!}{|\mathcal G|!} \Delta_G(S),
\end{equation}
which is a weighted average of marginal contributions to all possible combinations of groups.

The GSI cannot always be calculated in a time-efficient way, because the number of coalitions $S \subset \mathcal G \backslash G$ can become large very quickly. In practice, the Shapley value is often approximated \citep{casa2019, covert2020SAGE} by drawing $M \leq |\mathcal G|!$ different coalitions $S \subset \mathcal G \backslash G$ and averaging the marginal, weighted contributions:
\begin{equation}
    \label{eq:shapleygroupapprox}
    \hat{\phi}_{M}(G) = \frac{1}{M}\sum_{m = 1}^M (|\mathcal G| - 1 - |S_m|)! \cdot |S_m|! \cdot \Delta_G(S_m),
\end{equation}
with $S_m \subset \mathcal G \backslash G$, for all $m= 1, ..., M$.

While the GSI can be calculated with permutation- as well as refitting-based approaches, we will apply only the permutation-based approach in the upcoming simulation studies and the real-world example.

\subsubsection{Properties of the Grouped Shapley Importance}
For single features\footnote{Remember the one-to-one association of the numbers $1, ..., p$ and the features $\mathbf x_1,..., \mathbf x_p$} $x_i \in \{1,...,p\}$, which are divided into $l$ groups, we define the marginal contribution for $x_i$ as
$$
\Delta_{\{x_i\}}(S) = v(S\cup \{x_i\}) - v(S),
$$
for $S \subset \{1, ..., p\}\backslash \{x_i\}$. The Shapley importance for single features $\phi(x_i)$ can also be defined analogously to (\ref{eq:shapleygroupformula}). One interesting question is, does the GSI for a group $G\subset\{1, ..., p\}$ decompose into the sum of Shapley importances of features in $G$? In the following, we want to analyze the remainder
\begin{equation}
    \label{eq:shapleyR}
R = \phi(G) - \sum_{i \in G} \phi(x_i).
\end{equation}

Similar to the functional ANOVA decomposition \citep{hooker2007generalized}, we assume, that the value function for a coalition $S\subset\{1, ..., p\}$ can be broken down into main and interaction effects
\begin{equation}
v(S) = \sum_{x_i \in S} v(x_i) + \sum_{i \neq j}\epsilon_{ij} + \sum_{i\neq j \neq k} \epsilon_{ijk} + ... ,   
\label{eq:interaction}
\end{equation}
where $\epsilon_{i...m}$ is the effect of the interaction between the features $x_{i}, ..., x_{m} \in S$. Note, $v(G_1)$ cancels out, meaning that these interaction terms cannot be computed directly but are assumed to affect the ``payout'' of the value function. 

With the assumption in Eq. (\ref{eq:interaction}), it follows that the Shapley importance of a single feature $x_1$ (without loss of generality) can be written as
\begin{equation}
    \label{eq:shapleydecompsingle}
    \phi(x_1) = v(x_1) + \frac{1}{2} \left(\sum_{i \neq 1}^p \epsilon_{1i}\right)+ \frac{1}{3}\left(\sum_{i \neq j \neq 1}^p \epsilon_{1ij} \right) + ... + \frac{1}{p}\epsilon_{1...p}.
\end{equation}
The value function of the feature $x_1$ contributes to the Shapley importance with the weight $1$ and all possible interaction effects with feature $x_1$ contribute with the reciprocal length of the interaction effect.
We proved this assertion in Appendix \ref{sec:shapleyproof}. 
Similar to (\ref{eq:shapleydecompsingle}), the GSI of a group $G_1$ (w.l.o.g.) can be written as
\begin{equation}
    \label{eq:shapleydecompgroup}
    \phi(G_1) = v(G_1) + \frac{1}{2} \left(\sum_{i \neq 1}^k \epsilon_{G_1G_i}\right)+ \frac{1}{3}\left(\sum_{i \neq j \neq 1}^k \epsilon_{G_1G_iG_j} \right) + ... + \frac{1}{k}\epsilon_{G_1...G_k},
\end{equation}
where $\epsilon_{G_1...G_k}$ is the (non-computable) interaction effect between features of groups $G_1$, ..., $G_k$, where each group provides at least one feature. 
By using Eq. (\ref{eq:interaction}) on $v(G_1)$, we get:
\begin{align}
\label{eq:shapleyvG1}
v(G_1) &= \sum_{i \in G_1}v(x_i) + \sum_{i \neq j\in G_1}\epsilon_{ij} + \sum_{i\neq j \neq k\in G_1} \epsilon_{ijk} + ...
\end{align}
Looking back at Eq. (\ref{eq:shapleyR}), a lot of terms cancel out by using Eq. (\ref{eq:shapleydecompsingle}) and Eq. (\ref{eq:shapleyvG1}). The term $v(G_1)$, meaning all main effects $v(x_i), i \in G_1$, and all interaction effects $\epsilon_{i, ..., k}, 1\leq k \leq |G_1|$ between features within $G_1$, cancels out entirely. Furthermore, at least all two-way interaction effects between groups $\epsilon_{G_1G_i}, i = 2,...,k$ cancel out. 
A combination of higher-order interaction terms between features of $G_1$ and $\{1, ...,p\}\backslash G_1$ remain.\footnote{They mostly only partly cancel out, depending on the number of features within the groups $G_1, ..., G_k$.}
This means that the remainder $R$ is (usually) not equal to zero in case the applied algorithm learned a  higher-order  interaction between features of the regarded group and other groups. The higher the remainder, the larger the higher-order interaction effect. Thus, the remainder can be used as a quantification of learned higher-order interaction effects between features of different groups.

\subsection{Sequential Grouped Feature Importance}
In general, feature groups do not necessarily have to be distinct or independent of each other. 
When groups partly contain the same or highly correlated features, we may obtain high grouped feature importance scores for similar groups. 
This can lead to misleading conclusions regarding the importance of groups. 
Quantifying the importance of different combinations of groups is especially relevant in applications where extra costs are associated with using additional features from other data sources. 
In this case, one might be interested in the sparest, yet most important combination of groups or in understanding the interplay of different combinations of groups. 
Hence, in practical settings, it is often important to decide which additional group of features to make available (e.g., buy or implement) for modeling and how groups should be prioritized under economic considerations.

\cite{Gregorutti2015} introduced a method called \textit{grouped variable selection}, which is an adaptation of the recursive feature elimination algorithm from \cite{guyon2002gene} and uses permutation-based grouped feature importance scores for the selection of feature groups. 
In Algorithm \ref{alg:seqimp}, we introduce a sequential procedure which is based on the idea of stability selection \citep{meinshausen2010stability}. The procedure primarily aims at understanding the interplay of different combinations of groups by analyzing how the importance scores change after including other groups in a sequential manner.
We prefer a refitting-based over a permutation-based grouped feature importance score when the secondary goal is to find well-performing combinations of groups.
The basic idea is to start with an empty set of features and to sequentially add the next best group in terms of LOGI until no further substantial improvement can be achieved. 
Our sequential procedure is based on a greedy forward search and creates an implicit ranking by showing the order in which feature groups are added to the model. 
To account for the variability introduced by the model, we propose to use repeated subsampling or bootstrap with sufficient repetitions (e.g. 100 repetitions).
In Figure \ref{fig:gfssim} and \ref{fig:sankeyusecase}, we visualize the results in alluvial charts \citep{networkD3} to illustrate how we can gain further insights about the interplay of good combinations of groups of features using this sequential grouped feature importance procedure. It shows how frequently a group was selected given that another group was already included and thereby highlights robust combinations of groups.

Given a set of groups $\mathcal G = \{G_1, ..., G_k\}$, with $G_i \subset \{1, ..., p\}$, for $i = 1,...,k$, we are looking for a well-performing combination of groups $B = \bigcup_{i \in I} G_i \subset \{1, ..., p\}, I \subset \{1, ..., j\}$. 
Starting with an empty set $B= \emptyset$, the LOGI scores of each group $G_1$, ..., $G_k$ are assessed individually using an inner resampling strategy for LOGI where the data splits are the same for each group.
Without loss of generality, let $G_1$ be the best performing set of groups according to the mean LOGI score on the test datasets.
If the grouped feature importance score of $G_1$ exceeds a given threshold $\delta > 0$, we define $B = G_1$ and continue looking for a group to add. 
In other words, how well are the combinations $\{G_1, G_2\}, ..., \{G_1, G_k\}$ performing?
We define the LOGI score of sets of subsets as the LOGI score of the union of all subsets. 
Thus, we assess the LOGI scores of $G_1 \cup G_2, ..., G_1 \cup G_k$, and find the best performing combination of two groups, which contain the best working previous group. 
This procedure of iteratively adding a group is repeated until the performance threshold is no longer exceeded, yielding a well-working combination of groups, for example, $B = G_1 \cup G_3 \cup G_5$.


\begin{algorithm}[t]
  \SetKwData{Left}{left}\SetKwData{This}{this}\SetKwData{Up}{up}
  \SetKwFunction{Union}{Union}\SetKwFunction{FindCompress}{FindCompress}
  \SetKwInOut{Input}{input}\SetKwInOut{Output}{output}
  \Input{Set of groups $\mathcal G = \{G_1, ..., G_k\}$. \\ Improvement threshold $\delta >0$. \\ Number of repetitions for the outer data splitting.}
  \Output{For every outer data split: a good combination $B \subset \{1, ..., p\}$.}
  \caption{Sequential Grouped Feature Importance}
  \label{alg:seqimp}
  
  \For{Every outer data split}{
   
   Let $B = \emptyset$\;
    \For{$i = 1,...,k$}{
        \uIf{$i = 1$}{
          Define candidate set $\tilde{B}:= \left\{\tilde G \in \mathcal P(\mathcal G)\big| |\tilde G| = 1 \right\}$\;
          
          Find best working single group $G^* = \underset{\tilde G \in \tilde B}{\arg \max} \left(\widehat{LOGI}(\tilde G) \right)$;
          
           \uIf{$\widehat{LOGI}(G^*) > \delta$ }{
                $B = G^*$
                
                $L_{i-1} = \widehat{LOGI}(G^*) $
            }
        }
        \uIf{$i > 1$ and $B \neq \emptyset$}{
        Define candidate set $\tilde{B}:= \left\{\tilde G \in \mathcal P(\mathcal G) \big| |\tilde G| = i \text{ and } B \subset \tilde G \right\}$\;
        
                      Find best working combination $G^* = \underset{\tilde G \in \tilde B}{\arg \max} \left(\widehat{LOGI}\left( \bigcup_{G' \in \tilde G}G'\right) \right)$;
           
           \uIf{$\widehat{LOGI}\left( \bigcup_{G' \in G*}G'\right) - L_{i-1} > \delta$ }{
                $B = \bigcup_{G' \in G^*}G'$
                
                $L_{i-1} = \widehat{LOGI}\left( \bigcup_{G' \in G^*}G'\right)$
            }  
            \Else{
                \textbf{break} for loop
            }
        }
    }
  }
\end{algorithm}

    
\section{Comparison of Grouped Feature Importance Methods}
\label{sec:challenges}
After introducing the methodological background of the different loss-based grouped feature importance measures in Section \ref{sec:featimp}, we will now compare them in different simulation settings. 
We analyze the impact on all methods for settings where (1) groups are dependent, (2) correlations within groups vary, and (3) group sizes differ.

\subsection{Dependencies between Groups and Sparsity}
    \label{sec:sim_fi}
     
In this chapter, we compare refitting- and permutation-based grouped feature importance methods and show how different dependencies between groups can influence the importance scores. We demonstrate the benefits of the sequential grouped feature importance procedure and conclude with a recommendation when to use refitting or permutation-based methods depending on the use-case.    

We simulate a data matrix $\mathbf X$ with $n = 1000$ instances and $3$ groups $G_1, G_2, G_3$ with each of them containing $10$ normally distributed features. 
While features in $G_3$ are created such that they are almost uncorrelated with features of the other groups, $G_1$ and $G_2$ are highly dependent. 
For this purpose, we generate the 10 features of group $G_1$ based on a normally distributed prototype vector $\mathbf U \sim \mathcal N (0,1)$ as follows:
For every feature included in group $G_1$, we alter 10\% of the observations by adding a normally distributed error term $\epsilon \sim \mathcal N(0, 0.5)$ \citep[for a similar approach, see][]{Tolosi2011}. 
$G_2$ is generated by copying features of $G_1$ and adding a small normally distributed error term $\epsilon \sim \mathcal N(0, 0.01)$ to the copied features.
Features of the group $G_3$ are generated similar to group $G_1$ but using a prototype vector $\mathbf V$ which is independent of $\mathbf U$. 
The target vector $\mathbf Y$ is generated by $\mathbf Y = 2\mathbf U + 1\mathbf V + \epsilon$, with $\epsilon \sim \mathcal N(0, 0.1)$. 
We fitted a support vector machine with a radial basis function kernel\footnote{Epsilon regression, $\epsilon = 0.1, C=1$ with heuristically chosen kernel width according to \citep{caputo:2002} (here: $\sigma = 0.079$).}, as an example of a black-box algorithm.
    
The results in Table \ref{tab:sim} show that there can be major differences depending on how the grouped feature importance is calculated. 
Permutation methods (GOPFI \& GPFI \& GSI) reflect the importance of the groups based on a model trained on a fixed dataset.
In contrast, refitting methods (LOGI \& LOGO) retrain the model on a reduced dataset and can therefore learn new relationships. 
Looking at the results from the permutation methods, 
we can see, that the groups $G_1$ and $G_2$ share the same importance and are more important than $G_3$. 
The results from the refitting methods, however, can reveal some interesting relationships between the groups. 
The refitting methods highlight that $G_1$ and $G_2$ are more or less interchangeable, hence do not complement each other. 
This is reflected by the near-zero LOGO scores, which indicate, that leaving each group out of the full model does not change the model's expected loss considerably.

\begin{figure}
        \centering
        \includegraphics[width=0.7\textwidth]{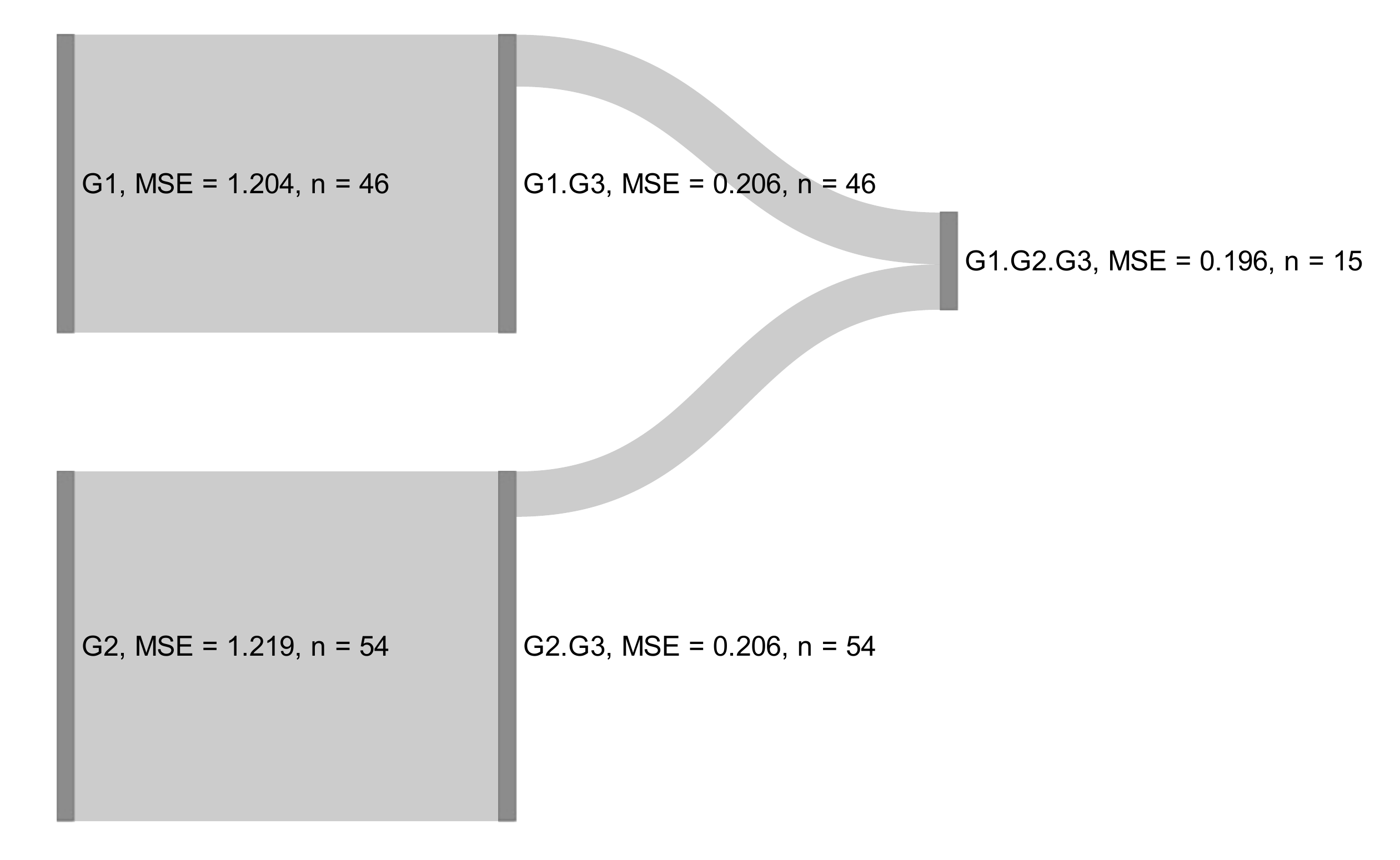}
        \caption{Sequential grouped feature importance for the simulation in Section \ref{sec:sim_fi}. 100 times repeated subsampling. Improvement threshold $\delta = 0.001$. Vertical bars show one step of the sequential procedure (left to right). Height of the vertical bars represent the number of subsampling iterations a combination of groups was chosen. $MSE$ scores show predictive performance. Streams represent the addition of a group.}
        \label{fig:gfssim}
    \end{figure}

Figure \ref{fig:gfssim} illustrates the results of the sequential procedure introduced in Algorithm \ref{alg:seqimp}. We see that across 100 subsampling iterations, $G_1$ was chosen 46 times as the most important first group, and $G_2$ was chosen 54 times with similar predictive performance for both groups.   
In the second step, the group $G_3$ was added in all cases to either $G_1$ or $G_2$. This step resulted in an on-average drop in the MSE score from 1.2 to 0.2.
Only in a few cases (15 out of 100), the final addition of either $G_1$ or $G_2$ to a full model was exceeding the very low chosen threshold of $\delta = 0.001$.
This reveals that these two groups are rather interchangeable and do not benefit from one another.
    
    \begin{table}[ht]
    \centering
\begin{tabular}{lrrrrr}
  \hline
  Group & GOPFI & GPFI & GSI & LOGI & LOGO \\ 
  \hline
  $G_1$ & 6.04 ($\pm$ 0.37) & 2.64 ($\pm$ 0.07)& 4.12 ($\pm$ 0.45)& 3.93 ($\pm$ 0.75) & -0.01 ($\pm$ 0.02) \\ 
   $G_2$ & 5.90 ($\pm$ 0.35) & 2.57 ($\pm$ 0.09)& 4.01 ($\pm$ 0.47)& 3.93 ($\pm$ 0.76)& -0.00 ($\pm$ 0.02) \\ 
   $G_3$ & 1.76 ($\pm$ 0.39)& 1.75 ($\pm$ 0.05)& 1.54 ($\pm$ 0.39)& 0.58 ($\pm$ 1.01)& 1.01 ($\pm$ 0.22)\\ 
   \hline
\end{tabular}
    \caption{Results of different feature importance calculations of the simulation. GSI scores were calculated without approximation with $v_\text{perm}$ as value function. All results were averaged by a 10-fold cross validation scheme with standard deviations reported in parentheses.}
    \label{tab:sim}
    \end{table}
    
The choice between using permutation or refitting grouped feature importance methods might depend on the number of groups and correlation strength between the different groups. If feature groups are distinct, and features between the groups are almost uncorrelated, we might prefer permutation over refitting methods due to lower computation time. In cases where groups are correlated with each other (e.g., because some features belong to multiple groups), refitting methods might be preferable as they are not misleading in correlated settings. Since the number of groups is usually smaller than the number of features in a dataset, refitting methods for groups of features could become a viable choice.
Furthermore, with the sequential grouped feature importance procedure it is possible to find sparse and good combinations of groups in an interpretable manner and thus helps to better understand dependencies and interactions between groups. 
    
\subsection{Varying Correlations within Groups}
In many use cases, it is quite common to group similar and therefore often correlated features together while groups of features may be almost independent of each other. However, compared to Section \ref{sec:sim_fi} correlations of features within groups might differ. 
We created a data matrix $\mathbf X$ with $n = 1000$ instances and $4$ groups $G_1$, $G_2$, $G_3$, $G_4$ with each of them containing $10$ normally distributed features.
Using 5-fold cross-validation, we fitted a random forest with 2000 trees and a support vector regression with a radial basis function kernel\footnote{We used a cost parameter of $C=1$ and estimate the kernel width based on the heuristic introduced by \citep{caputo:2002}}.
The univariate target vector $\mathbf Y$ is defined as follows:
\begin{align*}
		\mathbf Z_{j} &= 3\mathbf X_{G_j, 3}^2 - 4\mathbf X_{G_j, 5} - 6\mathbf X_{G_j, 7} + 5\mathbf X_{G_j, 9} \cdot d_j, \quad j \in \{1,2,3\}\\
		\mathbf Y &= \sum_{j=1}^3 \mathbf Z_j + \epsilon
\end{align*}
with 
\begin{equation*}
    d_j=
    \begin{cases}
      1, & \text{if}\ \text{mean}( \mathbf X_{G_j, 8}) > 0 \\
      0, & \text{otherwise}
    \end{cases}
  \end{equation*}
and $\epsilon \stackrel{iid}{\sim} N(0,1)$. The $i-$th feature of the $j$-th group is denoted by $\mathbf X_{G_j, i}$.
We repeated the simulation 20 times.

\begin{figure}
        \centering
        \includegraphics[width=0.98\textwidth]{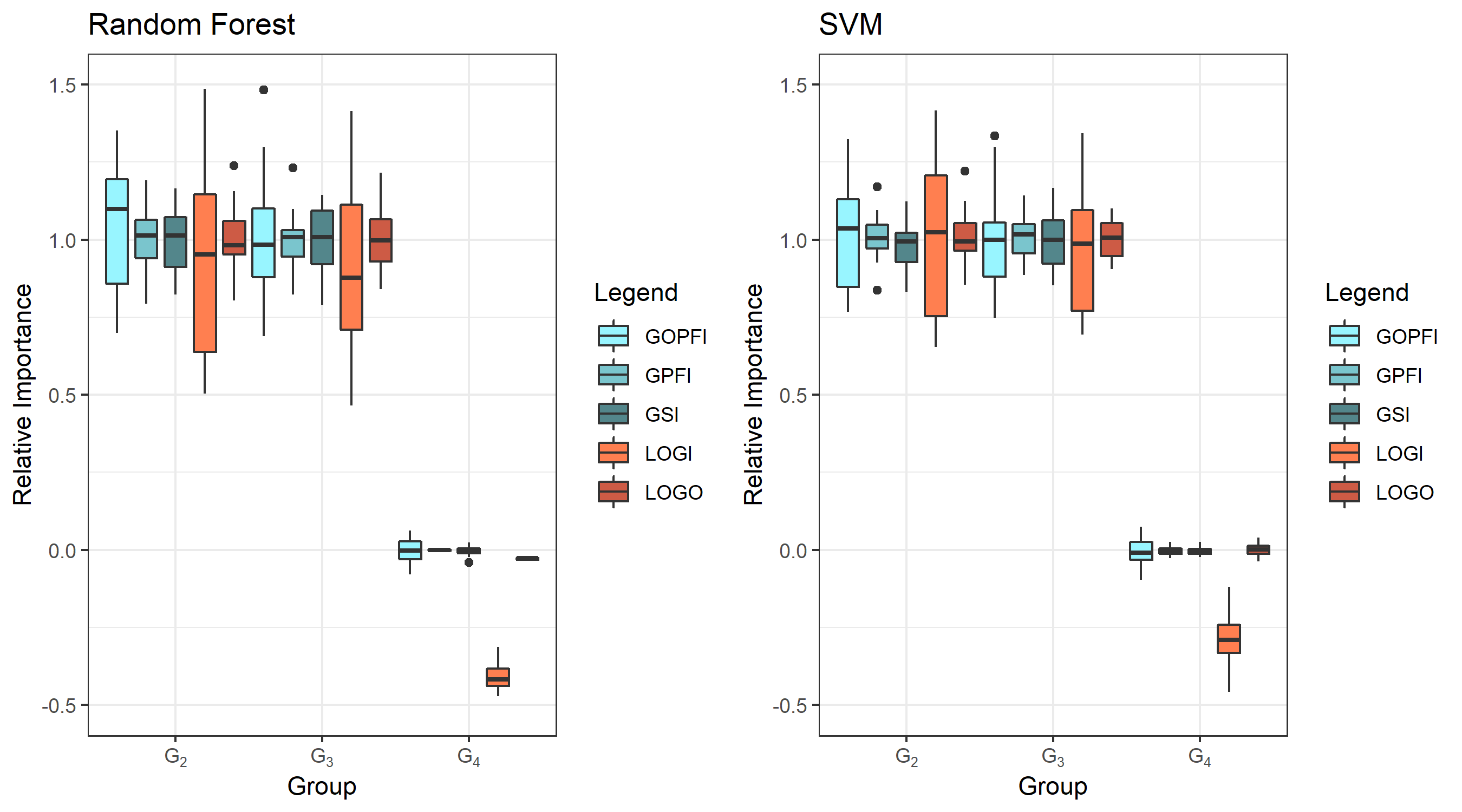}
        \caption{Grouped relative importance scores in case of equally sized within-group correlations for random forest (left) and SVM (right). Relative importance is calculated by dividing each of the absolute group importance scores by the importance score of $G_2$. Hence, relative importance of $G_1$ is 1. The boxplots illustrate the variation between different repetitions.}
        \label{fig:corr_same}
    \end{figure}

It follows that $G_1$, $G_2$, $G_3$ have the same influence on the target variable while $G_4$ has no influence on $\mathbf Y$.
Therefore, all features are generated from a prototype vector $\mathbf U$, which is sampled by a normal distribution $\mathcal N (0,1)$. 
For every feature, we alter a specific percentage of the observations by taking a weighted average between $\mathbf U$ ($20\%$) and an independent standard normally distributed random variable ($80\%$).
For the results shown in Figure \ref{fig:corr_same}, we set this percentage to $10\%$ for all features within the same group. Hence, correlations within groups are the same (around $90\%$) for all groups, while groups themselves are independent of each other. The plots show that all methods correctly attribute the same importance to the first three groups, while the fourth group being not important for predicting $\mathbf Y$. LOGI seems to be a bit less robust and can also take negative values in the case of group 4.

\begin{figure}
        \centering
       \includegraphics[width=0.98\textwidth]{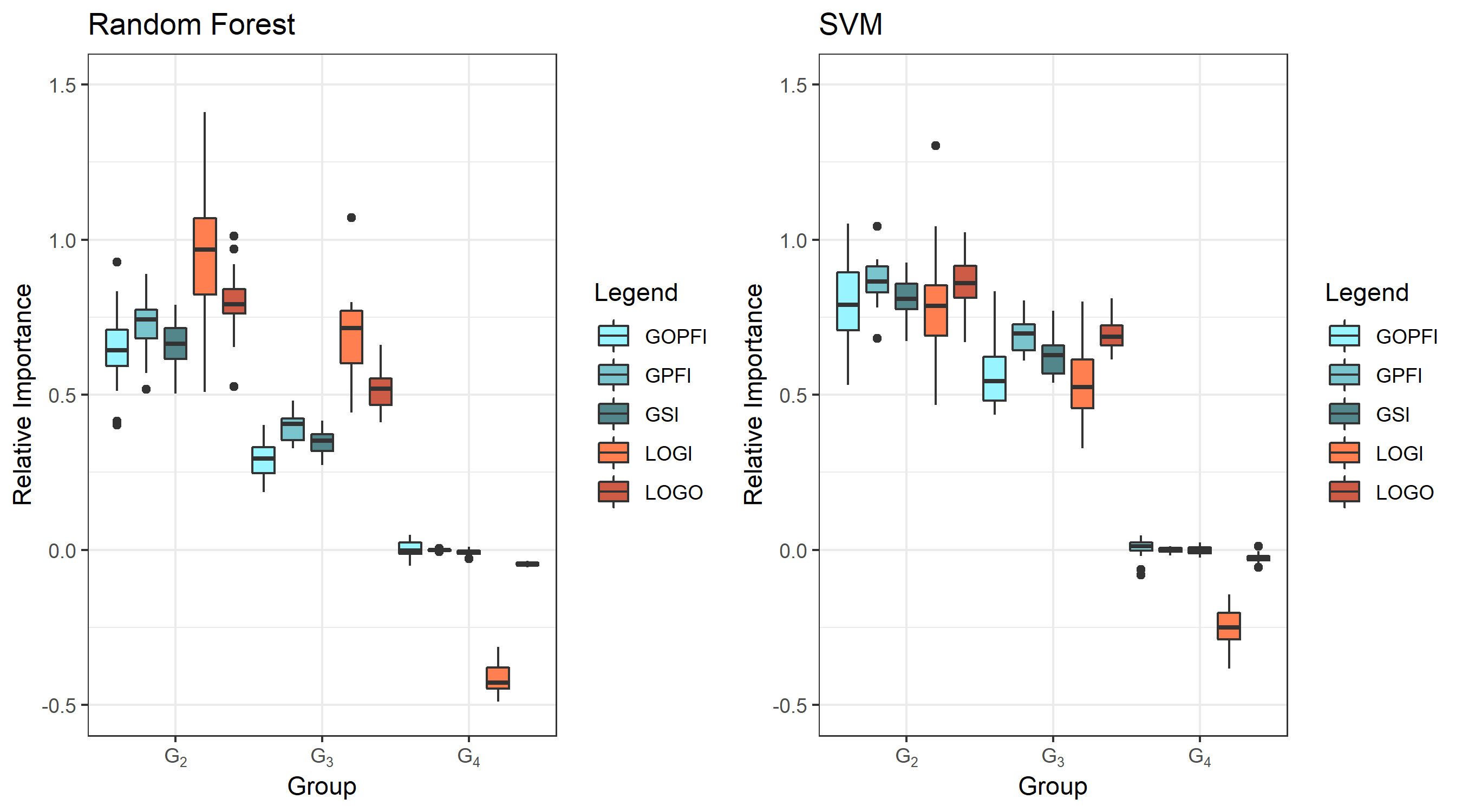}
        \caption{Grouped relative importance scores in case of varying sizes of within-group correlations for random forest (left) and SVM (right). Relative importance is calculated by dividing each of the absolute group importance scores by the importance score of $G_2$. Hence, relative importance of $G_1$ is 1. The boxplots illustrate the variation between different repetitions.}
         \label{fig:corr_diff}
    \end{figure}

\begin{figure}
        \centering
        \includegraphics[width=0.98\textwidth]{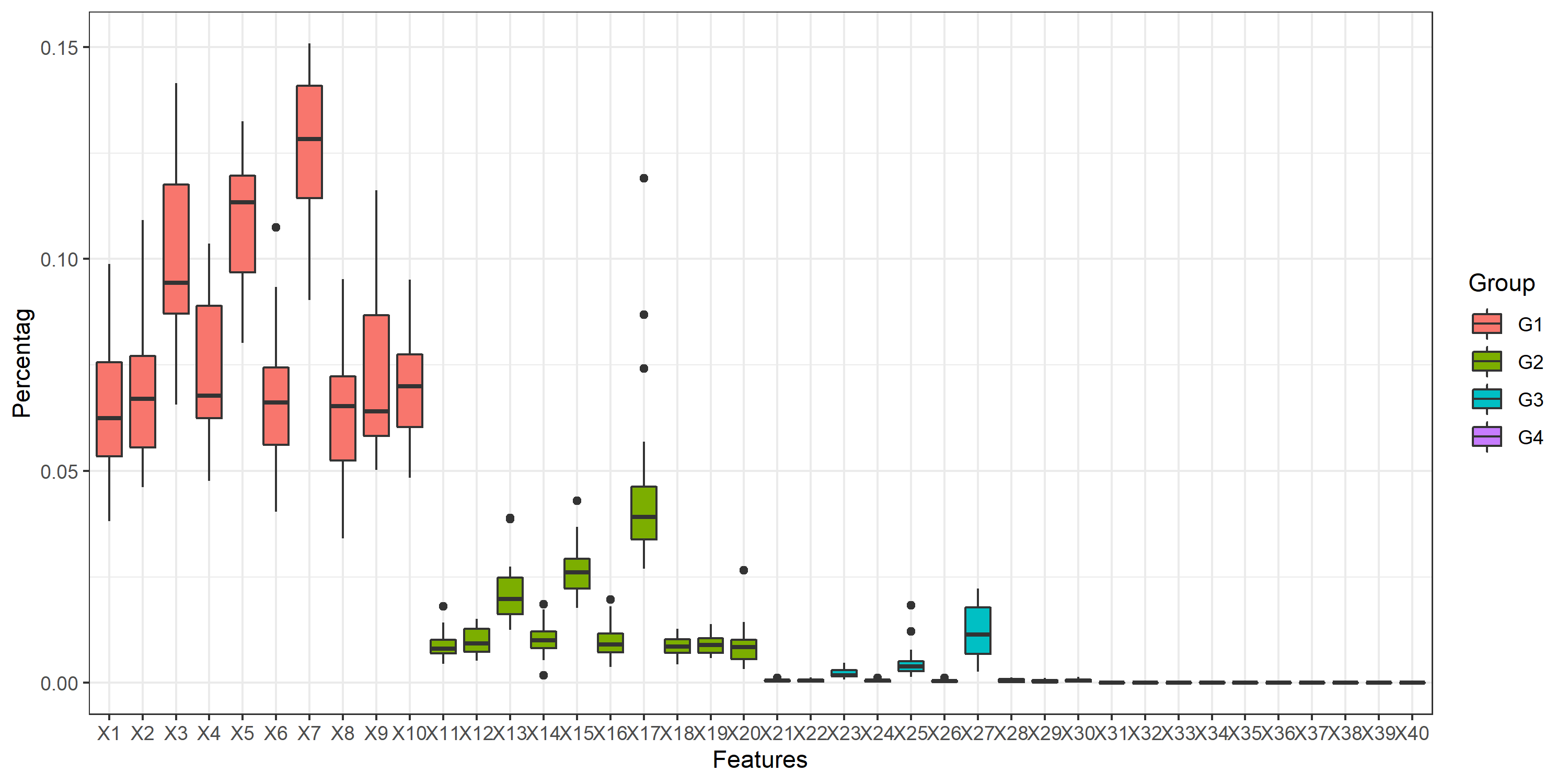}
        \caption{The figure shows the percentage of how often each feature is chosen as first splitting feature within the trained random forests. Results have been averaged over the cross validation folds for each repetition. The boxplots show the distribution over all 20 repetitions.}
         \label{fig:split_distr}
    \end{figure}

In Figure \ref{fig:corr_diff}, on the other hand, correlations within groups vary across groups. 
The altering percentage is set to $10\%$ for features of $G_1$ and $G_4$, to $30\%$ for features of $G_2$ and to $60\%$ for features of $G_3$. Hence, features in $G_1$ and $G_4$ are highly correlated within the respective group while features within $G_2$ and $G_3$ show a medium and small correlation, respectively. While $G_4$ is still recognized to be unimportant, the relative importance of groups 1 to 3 drops with decreasing within-group correlation. This artifact seems to be even more severe for the random forest compared to the support vector machine. For example, $G_3$ is on average less than half as important as $G_1$ for permutation-based methods.
Thus, none of the methods reflect the underlying true importance of the different groups. However, this might be due to the actually learned effects of different models, since grouped structures are not regarded in the modeling approach. 
Another possibility to quantify feature importance when using random forests is to extract the information on how often a feature has been used as a splitting variable for the different trees. The feature chosen for the first split has the most influence within each tree. Hence, we calculated for each repetition the percentage of how often a feature has been chosen as the first splitting feature. The distribution over all repetitions is displayed in Figure \ref{fig:split_distr}. Each of the features of $G_1$ is on average chosen more often as the first splitting feature than all features of the other groups, no matter if it has an influence on the target or not. The influential features of $G_3$ (which has the lowest within-group correlation) are barely chosen as the first splitting feature. 

Hence, users need to be careful in case there are varying dependency structures when using grouped feature importance methods on models that have been trained on single-feature space. Random forests are especially prone to bias in this case as shown in this simulation example as well as in other work such as \cite{Strobl2008}.

    
\subsection{Varying Sizes of Groups}
\begin{figure}
        \centering
        \includegraphics[width=0.9\textwidth]{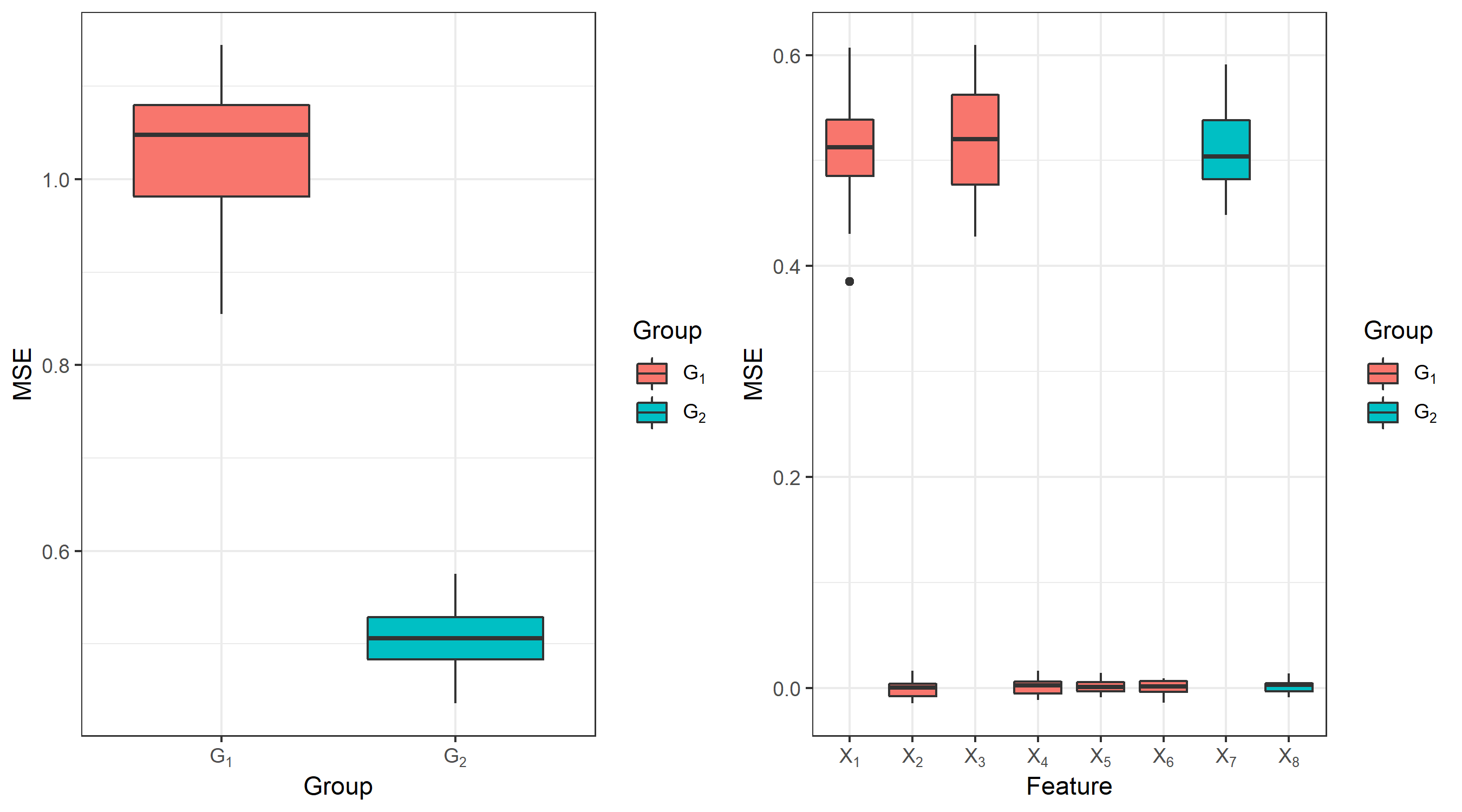}
        \caption{Shapley importance on group (left) and on feature level (right). Boxplots show the variation between the 20 repetitions of the experiment.}
        \label{fig:varying_size}
    \end{figure}

Another factor to consider when calculating grouped instead of individual feature importance scores is that differing group sizes might influence the ranking of the scores. 
Groups with more features might often have higher grouped importance scores and might contain more noise features than smaller groups. Therefore, \citet{Gregorutti2015} argue that in case one needs to decide between two groups that have an equal importance score, one would prefer the group with fewer features. Following from that, they normalize the grouped feature importance scores regarding the group size with the factor $|G|^{-1}$. This is also used in the default definition of the grouped model reliance score in \citet{valentin2020interpreting}. However, the usefulness of normalization highly depends on the question the user would like to answer. This is illustrated in a simulation example in Figure \ref{fig:varying_size}. 
We created a data matrix $\mathbf X$ with $n = 2000$ instances and $2$ groups with $G_1$ containing $\{x_1, \dots x_6\}$ and $G_2$ containing $\{x_7, x_8\}$ i.i.d. uniformly distributed features on the interval $[0, 1]$.
The univariate target variable $\mathbf Y$ is defined as follows:
\begin{align*}
		\mathbf Y &= 2\mathbf X_1 + 2\mathbf X_3 + 2\mathbf X_7 + \epsilon, \quad \text{with} \quad \epsilon \stackrel{iid}{\sim} N(0,1).
\end{align*}

We used $1000$ observations for fitting a random forest with $2000$ trees and $1000$ observations for prediction and calculating the GSI as defined in Section \ref{sec:shapleyImp} with a permutation-based value function. This was repeated $20$ times. Figure \ref{fig:varying_size} shows that $G_1$ is about twice as important as $G_2$. As shown in Section \ref{sec:shapleyImp} we can compare the GSI with the Shapley importance on feature level. In case there are no higher-order interaction terms between groups modeled by the random forest, the single feature importance scores will approximately sum up to the grouped importance score as shown in this example. This provides a more detailed view of how many and which features have been important within each group. In this case, there are two equally important features in $G_1$ and one equally important feature in $G_2$. If we use the normalization constant in this example, we would divide the grouped importance score of $G_1$ by 6 and the one of $G_2$ by 2, and hence $G_2$ would be regarded as more important than $G_2$. It follows that if we need to decide between two groups, we would choose $G_2$ when we follow the approach of \citet{Gregorutti2015} although the user might prefer $G_1$ since there are two features with the same importance as the one of $G_2$ and hence $G_1$ contains more information. Furthermore, breaking down the GSI to the single feature Shapley importance scores puts the user in the position to define sparser groups by excluding non-influential features.

\section{Feature Effects for Groups}
\label{sec:groupedEffect}
Feature effect methods quantify or visualize the influence of features on the model's prediction.
For a linear regression model, we can easily summarize the feature effect in one number making interpretation very simple: If we change feature $x_1$ by 1 unit, our prediction will change by the corresponding coefficient estimate $\hat \beta_1$ (positively or negatively depending on the sign of the coefficient). 
For more complex non-linear models like generalized additive models, such a simplified summary of the feature effect is not adequate since the magnitude and sign of the effect might change over the feature's value range. Hence, it is more common to visualize the marginal effect of the feature of interest on the predicted outcome. 
Since ML models are often complex non-linear models, different visualization techniques for the feature effect have been introduced in recent years. Common methods are 
PDP, ICE curves or ALE \citep{friedman2001greedy, icecurves, apley2016visualizing}, which show how changes in the feature values affect the predictions of the model. However, these are usually only defined for a maximum of two features. 
For larger groups of features, this becomes more challenging since it is difficult to simultaneously visualize the influence of several features. 
The approach described in this section aims to create effect plots for a predefined group of features that have a similar interpretation to the single-feature PDP. 
To achieve this, we transform the high-dimensional space of the feature group into a low-dimensional space by using a supervised dimension reduction method which is discussed in Section \ref{sec:dimred}. 
We want to find a few underlying factors that are attributed to a sparse and interpretable combination of features that explain the effect of the regarded group on the model's expected loss. We provide a detailed description of this method in Section \ref{sec:lpdp} and introduce the resulting CFEP. 
In Section \ref{sec:sim_dim}, we illustrate the advantages of applying a supervised instead of an unsupervised dimension reduction method and compare our method to the totalvis effect plot introduced in \cite{seedorff2021totalvis}.

\subsection{Choice of Dimension Reduction Method}
\label{sec:dimred}
The probably most prominent dimension reduction technique is the principal component analysis (PCA). PCA finds a projection $\mathbf V \in \mathbb R^{p \times p}$, which maximizes the total variance of projected data $\mathbf X\mathbf V$ through an Eigen decomposition of the sample covariance matrix.
PCA is restricted to explaining most of the variance of the feature space and the identified projections are not related to the target variable. 
Because we want to visualize the mean prediction of combined features as a result of the dimension reduction process, we prefer supervised procedures that maximize dependencies between the projected data $\mathbf X\mathbf V$ and the target vector $\mathbf Y$ (as we show in Section \ref{sec:sim_dim}). Many methods for supervised PCA (SPCA) have been established, see for example \citet{bair2006}, who used a subset of features that were selected based on their linear correlation with the target variable. Another very popular method that maximizes the covariance between features and the target variable is partial least squares (PLS) \citep{Wold1984}. The main difference of these methods compared to the SPCA introduced by \citet{barshan2011supervised} is that the SPCA is based on a more general measure of dependence, called the Hilbert-Schmidt Independence Criterion (HSIC). 
This independence measure is constructed to be zero, if and only if any bounded continuous function between the feature and target space is uncorrelated. In practice, an empirical version of the HSIC criterion is calculated with kernel matrices.
It follows that while this SPCA technique can cover all kinds of linear and non-linear dependencies between $\mathbf X$ and $\mathbf Y$ by choosing an appropriate kernel, the other suggested methods are only able to model linear dependencies between the features and the target variable. Probably best suited for our application of finding \textit{interpretable} sets of features in a high-dimensional dataset is the method called sparse SPCA, described in \cite{sharifzadeh2017sparse}. 
Similar to the SPCA method from \cite{barshan2011supervised}, sparse SPCA uses the HSIC criterion to maximize the dependency between projected data $\mathbf X\mathbf V$ and the target $\mathbf Y$ but also incorporates a $L_1$ penalty of the projection $\mathbf V$ for sparsity.
The sparse SPCA problem can be solved with a \textit{penalized matrix decomposition} \citep{witten2009penalized}. More theoretical details on the sparse SPCA, including the HSIC criterion and how it can be calculated empirically, and the choice of kernels and hyperparameters can be found in Appendix \ref{sec:sspca_details}.

\subsection{Totalvis Effect Plot}
\citet{seedorff2021totalvis} recently introduced a method that aims to plot the combined effect of multiple features by using PCA. Their approach can be described as follows: First, they apply PCA on the regarded feature space to receive the principal components matrix after rotation. For the principal component of interest, they create an equidistant grid. Second, for each grid value, they replace all values of the selected principal component with this grid value and transform the matrix back to the original feature space. Third, The ML model is applied on these feature values and a mean prediction for the grid point of the regarded principal component is calculated. Steps 2 and 3 are repeated for all grid points. 

Hence, with this method combined effect plots for up to $p$ principal components can be created. Thus, \citet{seedorff2021totalvis} do not focus on explaining groups of features explicitly. Furthermore, they use PCA for dimension reduction which is unsupervised, and thus projections might not be related to the target. Since using PCA and not sparse PCA, the results might be hard to interpret since many or all features might have an influence on the principal component. Last but not least, with the back-transformation from the principal component matrix to the original feature space, all feature values change and might not be meaningful anymore. For example, in the case of integer features, the back-transformation might lead to real feature values.  
We illustrate the disadvantages of the method compared to the CFEP in Section \ref{sec:sim_dim}.

\subsection{Combined Features Effect Plot}
\label{sec:lpdp}
\begin{figure}[t]
    \centering
    \begin{subfigure}{.7\textwidth}
        \begin{eqnarray*}
        \centering
         \scalebox{.55}{
        \begin{tabular}{|ccc|cc|}
          \hline
        \multicolumn{3}{|c|}{Group 1} & \multicolumn{2}{|c|}{Group 2}\\
        \hline
        $\mathbf x_1$ & $\mathbf x_2$ & $\mathbf x_3$ & $\mathbf x_4$ & $\mathbf x_5$ \\ 
          \hline
        \cellcolor{red!30} 1  & \cellcolor{red!30} -1  & \cellcolor{red!30} 2  & \cellcolor{black!30} 2.5  & \cellcolor{black!30} 3  \\ 
          \cellcolor{green!30} -2  & \cellcolor{green!30} 1.5  & \cellcolor{green!30} 3  & \cellcolor{black!30} -2  & \cellcolor{black!30} -1  \\ 
          \cellcolor{blue!30} 2.3  & \cellcolor{blue!30} 4  & \cellcolor{blue!30} -1  & \cellcolor{black!30} 6  & \cellcolor{black!30} 2  \\ 
          \cellcolor{yellow!30} -6.5  & \cellcolor{yellow!30} 8  & \cellcolor{yellow!30} 0  & \cellcolor{black!30} 5  & \cellcolor{black!30} 1  \\ 
          \cellcolor{magenta!30} 0.5  & \cellcolor{magenta!30} 1  & \cellcolor{magenta!30} 2  & \cellcolor{black!30} 4  & \cellcolor{black!30} 2  \\ 
          \cellcolor{cyan!30} 4  & \cellcolor{cyan!30} -2  & \cellcolor{cyan!30} 2  & \cellcolor{black!30} 3  & \cellcolor{black!30} 3  \\ 
           \hline
        \end{tabular}
        }
        \rightarrow
        \scalebox{.55}{
        \begin{tabular}{|ccc|cc|}
          \hline
          \multicolumn{3}{|c|}{Group 1} & \multicolumn{2}{|c|}{Group 2}\\
        \hline
        $x_1^{(1)}$ & $x_2^{(1)}$ & $x_3^{(1)}$ & $\mathbf x_4$ & $\mathbf x_5$ \\ 
          \hline
        \cellcolor{red!30} 1  & \cellcolor{red!30} -1  & \cellcolor{red!30} 2  & \cellcolor{black!30} 2.5  & \cellcolor{black!30} 3  \\ 
          \cellcolor{red!30} 1  & \cellcolor{red!30} -1  & \cellcolor{red!30} 2 & \cellcolor{black!30} -2  & \cellcolor{black!30} -1  \\ 
          \cellcolor{red!30} 1  & \cellcolor{red!30} -1  & \cellcolor{red!30} 2  & \cellcolor{black!30} 6  & \cellcolor{black!30} 2  \\ 
          \cellcolor{red!30} 1  & \cellcolor{red!30} -1  & \cellcolor{red!30} 2  & \cellcolor{black!30} 5  & \cellcolor{black!30} 1  \\ 
          \cellcolor{red!30} 1  & \cellcolor{red!30} -1  & \cellcolor{red!30} 2  & \cellcolor{black!30} 4  & \cellcolor{black!30} 2  \\ 
          \cellcolor{red!30} 1  & \cellcolor{red!30} -1  & \cellcolor{red!30} 2  & \cellcolor{black!30} 3  & \cellcolor{black!30} 3  \\ 
           \hline
        \end{tabular}
        }
        \underset{\text{predict}}{\rightarrow}
        \scalebox{0.55}{
        \begin{tabular}{|c|}
          \hline\\
        $\hat{\mathbf{y}}^{(1)}_{rep}$ \\ 
          \hline
        \cellcolor{black!30} 0.8  \\ 
          \cellcolor{black!30} 0.2  \\ 
          \cellcolor{black!30} 0.7  \\ 
          \cellcolor{black!30} 0.6  \\ 
          \cellcolor{black!30} 0.4  \\ 
          \cellcolor{black!30} 0.3  \\ 
           \hline
        \end{tabular}
        }
        \end{eqnarray*}
    \end{subfigure}\\
    \vspace*{0.3cm}
    \begin{subfigure}{.8\textwidth}
        \centering
          \includegraphics[width=0.8\textwidth]{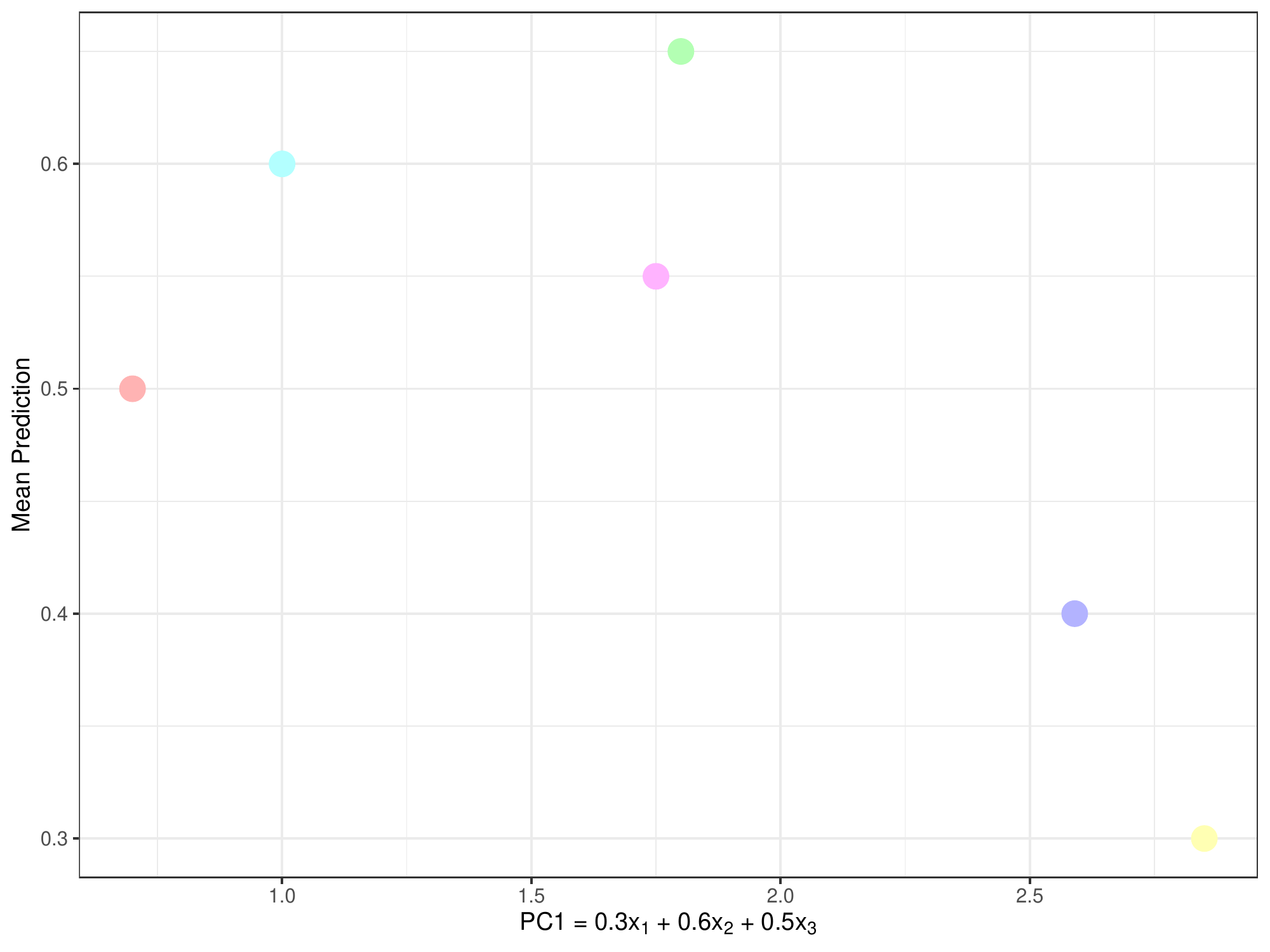}
    \end{subfigure}
    \caption{Explanation of estimating and visualizing CFEP; the x-coordinate reflects the linear combination of features with non-zero loadings for PC1 and the y-coordinate the mean predictions $\bm\bar{\hat{y}}^{(i)}_{rep}$ for each observation $i$.}
    \label{fig:CFEP_expl}
\end{figure}

To construct a CFEP for a defined group of features we first need to apply a dimension reduction method on this feature group to create a low-dimensional representation. In the case of sparse SPCA, we can obtain a reasonable number of influential features for each principal component. The CFEP illustrates the mean predictions for the sparse combination of features on observation level. The estimation of these mean predictions is explained in Figure~\ref{fig:CFEP_expl}. 
In this illustrative example, we have two predefined groups of features where the first group contains $x_1$, $x_2$ and $x_3$ and features $x_4$ and $x_5$ belong to the second group. 
To calculate the (grouped) mean prediction for the first group and first observation (shown in red), we replace the values of each instance in the dataset for the first group by the values of the first observation and predict $\hat{\mathbf{y}}^{(1)}_{rep}$ with the previously trained model. The value on the y-axis for the red point in the graph below corresponds to the mean over all predictions for the first observation: $\bm\bar{\hat{y}}^{(1)}_{rep}$ = $(0.8 + 0.2 + 0.7 + 0.6 + 0.4 + 0.3) / 6 = 0.5$. The value on the x-axis is the linear projection of the first observation for the regarded principal component. Hence, it is calculated by the weighted sum of feature values $x_1^{(i)}$ to $x_3^{(i)}$ where the weights are defined by the loadings of the respective principal component that we receive with sparse SPCA.
Hence, this method is computationally cheaper than totalvis, since we do not need any back transformations to the original feature space and we only need to calculate predictions once for each group of features and not for every principal component.

In contrast to PDP or totalvis effect plots we receive a point cloud instead of a curve.
The CFEP is, mathematically speaking, not a function, since points on the x-axis correspond to linear projections from a group of features. A point $z$ on the x-axis can have multiple combinations of features, which lead to $z$ and have different mean predictions on the y-axis.
However, we have now the possibility to interpret the shape of the point cloud and can draw conclusions about the behavior of the mean prediction of the model regarding the principal component.

This procedure is defined in a more general way in Algorithm~\ref{algo}. For this, let $G \subset \{1, ..., p\}$ be a group of features with $G = \{i_1,...,i_k \}$ and 
$\mathcal X_{G} := \mathcal X_{i_1} \times ... \times \mathcal X_{i_k}$.
A dimension reduction is a function $g: \mathcal X_{G} \longrightarrow \mathbb R$ that can be reasonably interpreted, like a linear projection.

\begin{algorithm}[t]
  \SetKwData{Left}{left}\SetKwData{This}{this}\SetKwData{Up}{up}
  \SetKwFunction{Union}{Union}\SetKwFunction{FindCompress}{FindCompress}
  \SetKwInOut{Input}{input}\SetKwInOut{Output}{output}
  \Input{Dataset $\mathcal D = \{(\mathbf x^{(i)}, y^{(i)})\}_{i = 1}^n$, group-defining subset $G \subset \{1,...,p\}$, dimension reduction function $g: \mathcal X_{G} \longrightarrow \mathbb R$, model $\hat{f}$ trained on $\mathcal D$.}
  \Output{Combined Features Effect Plot}
  \For{$i \in \{1, ..., n\}$}{
   get feature values $(\mathbf x_j^{(i)})_{j \in G}$\;
   create $\mathcal D_{rep}^{(i)}$ by replacing feature values of every other observation with these feature values\;
   predict $\hat{\mathbf{y}}^{(i)}_{rep}$ by applying $\hat{f}$ on $\mathcal D_{rep}^{(i)}$\;
   calculate the mean prediction $\bm\bar{\hat{y}}^{(i)}_{rep}$ of $\hat{\mathbf{y}}^{(i)}_{rep}$\;
   save $[g((\mathbf x_j^{(i)})_{j \in G})$, $\bm\bar{\hat{y}}^{(i)}_{rep}$] as x- and y-coordinates of observation $i$ for the CFEP\;
  }
  \caption{Combined Features Effect Plot}
\label{algo}
\end{algorithm}



\FloatBarrier
\subsection{Experiments on Supervised vs. Unsupervised Dimension Reduction}
\label{sec:sim_dim}
As discussed in Section \ref{sec:dimred}, PCA might be the most popular method regarding dimension reduction and thus for example used in \citet{seedorff2021totalvis} in a related approach.
However, since PCA is unsupervised, it does not account for the dependency between feature space and the target variable.
To evaluate how much this drawback influences CFEP, we look at two regression problems on simulated data.
The first is defined by a single underlying factor depending on a sparse set of features, which can be represented by a single principal component. 
The linear combination of this feature set is also linearly correlated with the target variable. 
The second regression problem contains two underlying factors depending on two sparse sets of features.
While the linear combination of the first feature set is also linearly correlated with the target, the second factor has a quadratic effect on $\mathbf Y$. 
In both cases, we compare the usage of sparse supervised and unsupervised PCA (sparse SPCA and sparse PCA) as dimension reduction methods within CFEP and compare it to the totalvis effect plot. 
Here, we investigate if the respective dimension reduction method does correctly identify the sparse set of features for each group.
Additionally, we determine how accurately we can predict the true underlying relationship between the linear combination of these features and the target variable.
Since we simulated the data, we know the number of underlying factors (principal components).

\subsubsection{One Factor}
In this example, we created a data matrix $\mathbf X$ with 500 instances of 50 standard normally distributed features with decreasing correlations. 
Therefore, all features are generated from a prototype vector $U$, which is sampled by a normal distribution $\mathcal N (0,1)$. 
For every feature, we alter a specific percentage of the observations by taking a weighted average between $\mathbf U$ ($20\%$) and an independent standard normally distributed random variable ($80\%$).
This percentage is set to $20\%$ for the first 10 features, to $40\%$ for the next 10 features up to $100\%$ for the last ten features.
Thus, while the first ten features are highly correlated with each other, the last ten features are almost uncorrelated.
The sparse subgroup defined by the variable $\mathbf Z$ is a linear combination of five features from $\mathbf X$ and has itself a linear effect on the univariate target variable $\mathbf Y$:
\begin{align*}
		\mathbf Z &= \mathbf X_5 - 2\mathbf X_8 - 4\mathbf X_{25} + 8\mathbf X_{47} + 4\mathbf X_{49}\\
		\mathbf Y &= \mathbf Z + \epsilon, \quad \text{with} \quad \epsilon \stackrel{iid}{\sim} N(0,1).
\end{align*}
Hence, according to our notation, $G_{\mathbf Z}$ is defined by $G_{\mathbf Z} = \{5,8,25,47,49\}$ and thus $X_{G_Z}$ is the related subset of all features. 
We drew 100 samples and fitted each time a random forest with 2000 trees. We used the 10-fold cross-validated results to perform sparse SPCA. 
For each dimension reduction method, we estimated $\hat{\mathbf Z}$ by summing up the (sparse) loading vector (estimated by the dimension reduction method) multiplied by the feature matrix $\mathbf X$. Therefore, $\mathbf X_{G_{\hat{\mathbf Z}}}$ is defined by the received sparse feature set.
The mean prediction $\bm\bar{\hat{\mathbf Y}}_{rep}$ for the CFEP was calculated as described in Section~\ref{sec:lpdp}.

\begin{figure}
    \centering
    \includegraphics[width=0.98\textwidth]{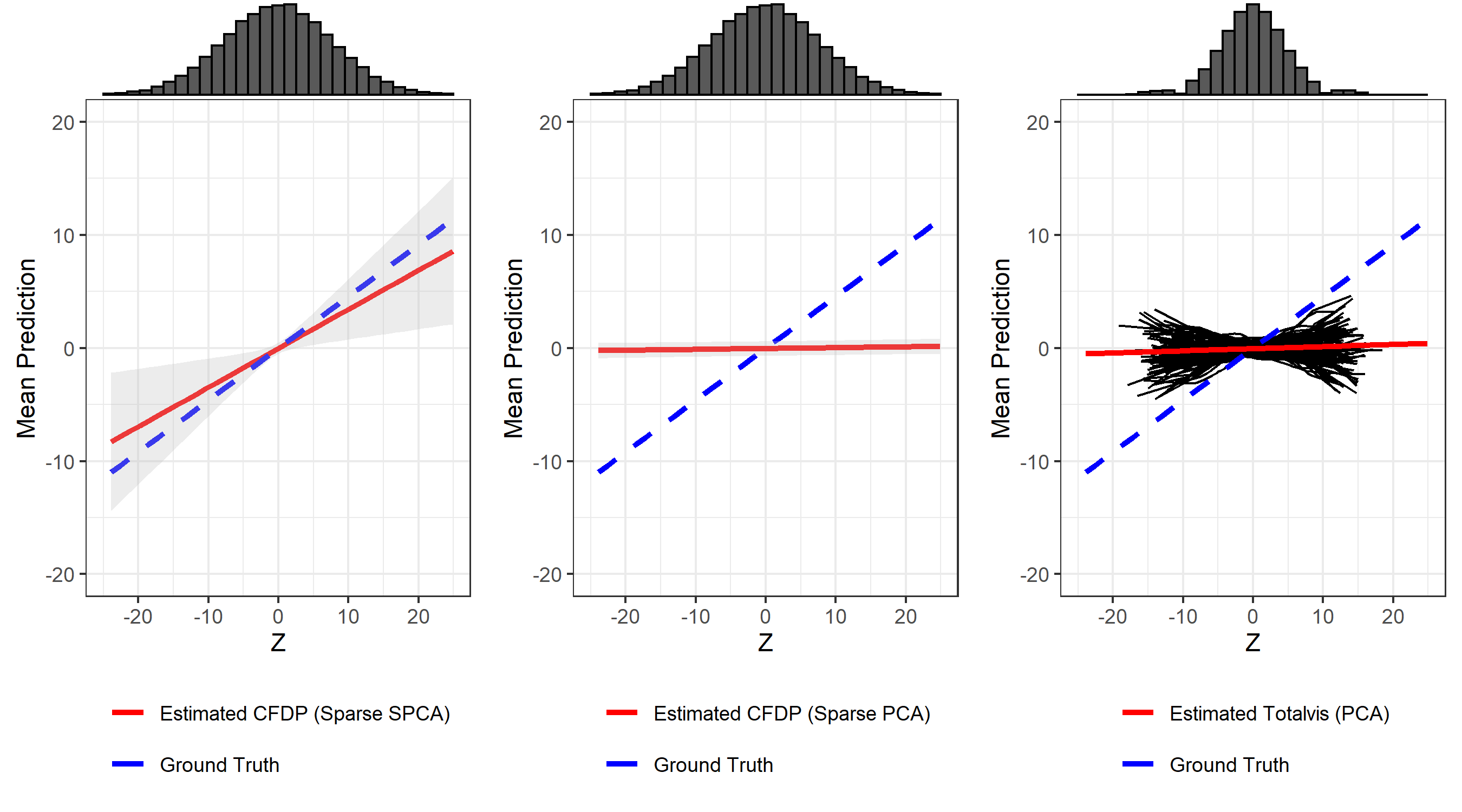}
    \caption{Average linear trend and confidence bands of CFEP over all samples using sparse SPCA (left) and sparse PCA (middle) compared to estimated totalvis effect curves over all 100 samples for first principal component (black) and the average linear trend (red) (right).}
    \label{fig:pd_linear}
\end{figure}

\begin{figure}
    \centering
    \includegraphics[width=0.98\textwidth]{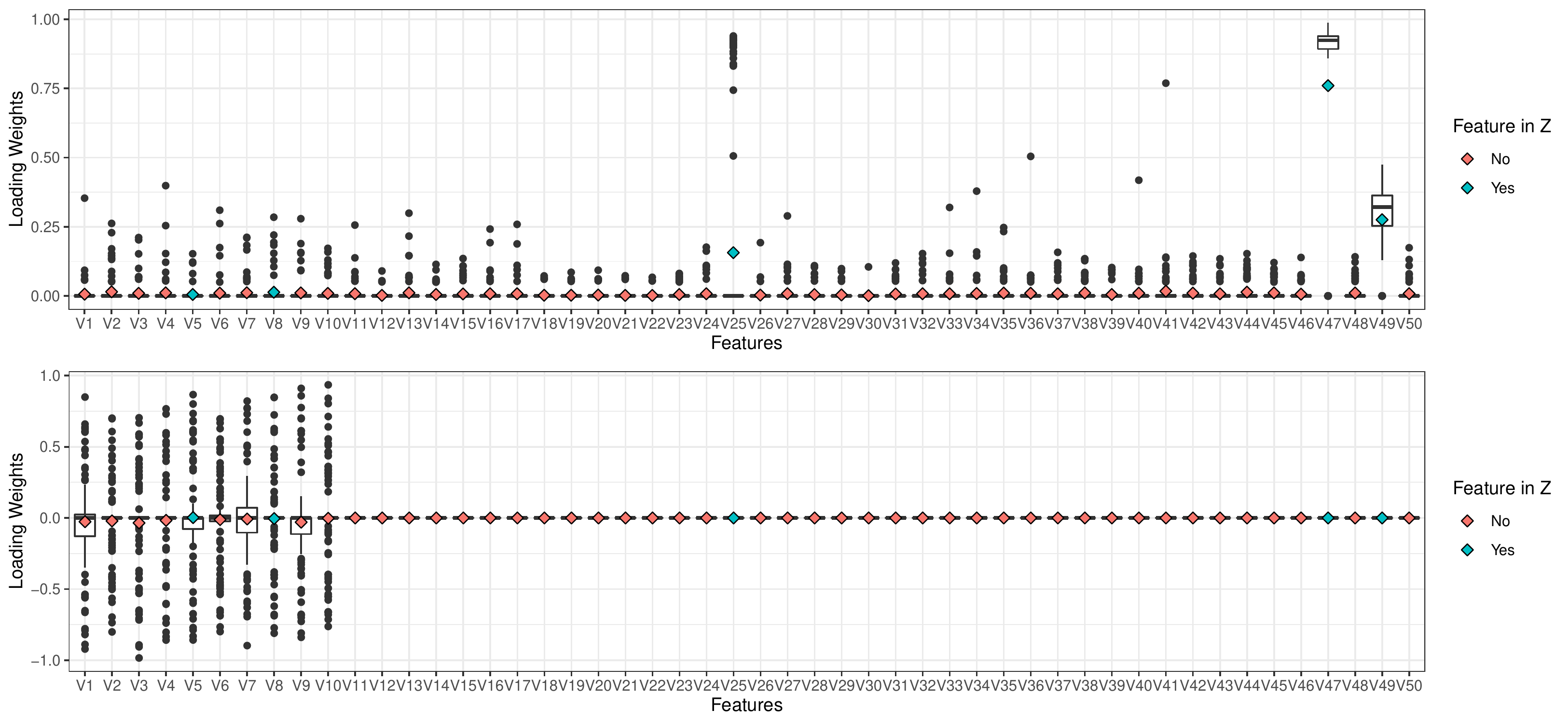}
    \caption{Distribution of feature loadings in sparse SPCA (top) and sparse PCA (bottom) over all samples; the rhombuses denote the mean values with the blue ones indicating the features that have an influence on the target in the underlying model formula.}
    \label{fig:feat_linear}
\end{figure}

The impact of choosing a supervised over an unsupervised sparse PCA approach is shown in Figure~\ref{fig:pd_linear}, which also shows the average linear trend and $95\%$ confidence bands of CFEP for the simulation results. 
To evaluate how well the estimated mean prediction $\bm\bar{\hat{\mathbf Y}}_{rep}$ approximates the underlying trend, we assume that we know that $\mathbf Z$ has a linear influence on the target. Thus, we fit on each simulation result a linear model. To compare the received regression lines, we evaluate each of them on a predefined grid and average over all 100 samples (represented by the red line). The confidence bands are then calculated by taking the standard deviation over all estimated regression lines on grid level and calculating the $2.5\%$ and $97.5\%$ quantiles using the standard normal approximation.
The associated calculation steps for each of the 100 samples can be summarized as follows:
\begin{itemize}
 \setlength{\itemsep}{0pt}\setlength{\parskip}{0pt}
    \item[1)] Estimate a linear model $\hat{f}(\mathbf \mathbf X_{G_{\hat{\mathbf Z}}}) \sim \mathbf Z$.
    \item[2)] Define an equi-distant grid of length 50 within the range of $\mathbf Z$.
    \item[3)] Apply the linear model estimated in 1) on the grid defined in 2). 
    \item[4)] Repeat steps 1 to 3 for $\hat{f}(\mathbf X_{G_\mathbf {Z}})$ hence using the true underlying features of $\mathbf Z$ to calculate the combined features dependencies which we call the ground truth.
\end{itemize}
The left plot in Figure~\ref{fig:pd_linear} clearly shows a similar linear trend of the estimated CFEP compared to the average ground truth (represented by the blue line) while the red line in the right plot varies around 0. By using sparse SPCA, the underlying feature set $\mathbf X_{G_{\hat{\mathbf {Z}}}}$ is better approximated than with sparse PCA which is reflected in the MSE between $\mathbf Z$ and $\hat{\mathbf Z}$ of 0.7 for sparse SPCA and 1.9 for sparse PCA. Figure~\ref{fig:feat_linear} provides an explanation for those differences. While sparse SPCA puts on average higher weights on features that have a high influence on the target, impactful loading weights for sparse PCA are solely distributed over highly correlated features in $\mathbf X$ that explain the most variance in feature space. Thus, including the relationship between the target and $\mathbf X$ in the dimension reduction method may have a huge influence on correctly approximating the underlying factor and hence also on the CFEP. 

Similar to using sparse PCA as a dimension reduction method within CFEP, the totalvis effect curves based on PCA do not show a clear positive linear trend on average (see Figure \ref{fig:pd_linear}). For almost half of the samples, we even receive a negative instead of a positive trend for the underlying factor. Thus, the interpretation is opposite to the actual effect and hence misleading.

\subsubsection{Two Factors}
In real-world data, it is usually the case that we have more than one underlying factor and also non-linear relationships of those on the target. Hence, we are now looking at a more complex simulation setting to see if we can observe the same behavior that we observed for the simple case. Therefore, we simulated a data matrix $\mathbf X$ with 500 instances for two feature sets each containing 20 standard normally distributed features. The data for each feature set is generated as described in the one-factor example but with an altering proportion of $15\%$ and $35\%$ for the features in the first set and $55\%$ and $85\%$ in the second set. Hence, the first ten features of each set show a higher correlation among each other than the last ten features and features of the first set are on average higher correlated than the second set. Features between the two sets are uncorrelated. The first factor $\mathbf Z_1$ is a linear combination of four features from the first set and $\mathbf Z_2$ of two features from the second set. $\mathbf Z_1$ has a linear and $\mathbf Z_2$ a quadratic effect on $\mathbf Y$.
\begin{align*}
		\mathbf Z_1 &= 3\mathbf X_3 - 2\mathbf X_8 - 4\mathbf X_{13} + 8\mathbf X_{18}\\
        \mathbf Z_2 &= 2\mathbf X_{25} + 4\mathbf X_{35}\\
		\mathbf Y &= \mathbf Z_1 + \mathbf Z_2^2 + \epsilon, \quad \text{with} \quad \epsilon \stackrel{iid}{\sim} N(0,1).
\end{align*}	\\

Again we drew 100 samples and fitted each time a random forest with 2000 trees. The approach is the same as described for one factor with the difference that we use the first two principal components as we want to find two sparse feature sets instead of one.

\begin{figure}
    \centering
    \includegraphics[width=0.98\textwidth]{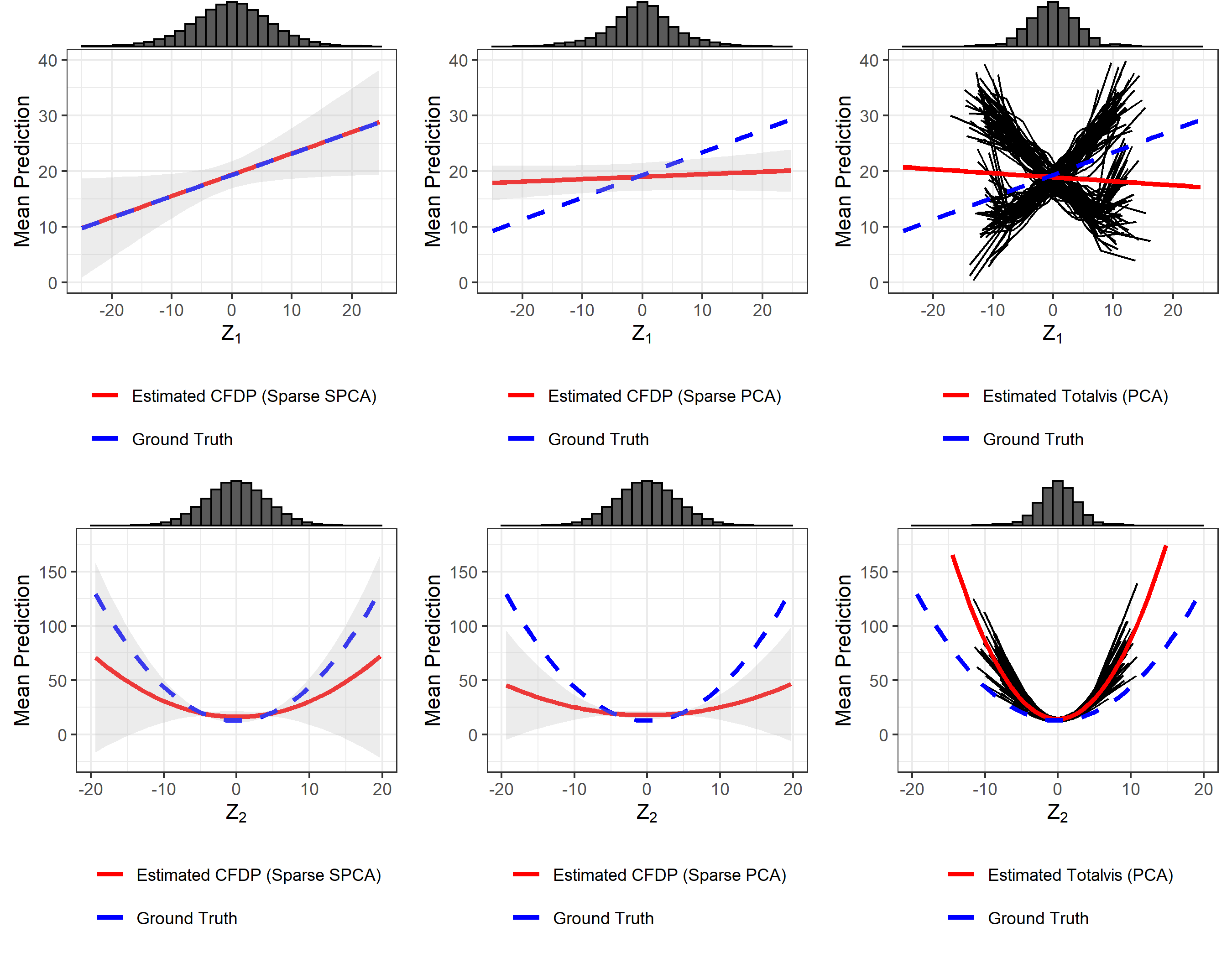}
    \caption{Top ($\mathbf Z_1$): Average linear trend and confidence bands of CFEP over all samples using sparse SPCA (left) and sparse PCA (middle) compared to estimated totalvis effect curves over all 100 samples for first principal component (black) and the average linear trend (red) (right). Bottom ($\mathbf Z_2$): Same structure as for $\mathbf Z_1$, but showing the quadratic trend of $\mathbf Z_2$.}
    \label{fig:pd_latent1}
\end{figure}

In Figure~\ref{fig:pd_latent1} the average linear and quadratic trend of the underlying CFEPs of $\mathbf Z_1$ and $\mathbf Z_2$ are depicted for both dimension reduction methods. While the average linear regression line of sparse SPCA matches the average ground truth almost perfectly for $\mathbf Z_1$, the associated line of sparse PCA shows only a very slightly positive trend and differs a lot from the average ground truth. A similar propensity can be observed for the quadratic shape regarding $\mathbf Z_2$. Again, this behavior is because sparse SPCA puts on average higher weights on features that have a high effect on the target, while the unsupervised version focuses on features that explain the most variance in $\mathbf X$.

The estimated linear trend of the totalvis effect curves for the first principal component is negative instead of positive and thus for most of the samples and on average completely misleading (see Figure \ref{fig:pd_latent1}). The quadratic shape of the second component is on average and for almost all samples steeper than the average ground truth. Also here, the deviation is higher than for CFEP with sparse SPCA.

\FloatBarrier

\FloatBarrier

\section{Real Data Example: Smartphone Sensor Data}
\label{sec:realdata}
 Smartphones and other consumer electronics have increasingly been used to collect data for research \citep{Miller2012, Raento2009}. The emerging popularity of these devices for data collection is grounded in their connectivity, the number of built-in sensors, and their widespread use. Moreover, smartphones are enabling users to perform a wide variety of activities (e.g., communication, shopping, dating, banking, navigation, listening to music) and thus provide an ideal way to study human behavior in naturalistic contexts, over extended periods of time, and at fine granularity \citep{Harari2015, Harari2016, Harari2017}. In this regard, smartphone data has been used to investigate individual differences in personality traits \citep{Stachl2017, Harari2019}, in human emotion and well-being \citep{ServiaRodriguez2017, Rachuri2010, Saeb2016, Thomee2018, Onnela2016}, and in day and nighttime activity patterns \citep{Schoedel2020}.
 
 We use a dataset on human behavior, collected with smartphones, to illustrate methods for group-based feature importance. The PhoneStudy dataset has been created from three separate datasets \citep{Stachl2017, Schuwerk2019, Schoedel2018}. It consists of 1821 features on smartphone-sensed behavior and 35 target variables on self-reported Big Five personality trait dimensions and subdimensions. The dataset has been published online and is openly available\footnote{\href{https://osf.io/kqjhr/}{https://osf.io/kqjhr/}}. The original study \citep{Stachl2020} used the behavioral variables to predict self-reported Big Five personality trait scores on 35 dimensions and explored which classes of behaviors were most predictive for each personality trait dimension and overall. The personality prediction task is challenging because, (1) the dataset contains many variables on similar behaviors, (2) these variables are often correlated, and (3) effects with the targets are interactive, very small, and partially non-linear. Many variables in the dataset can be manually grouped into classes of behavior (e.g., communication and social activity, app-usage, music consumption, overall phone activity, mobility).
 
We use this dataset to illustrate the idea of grouped feature importance with regard to the prediction of personality trait scores for the dimension of Conscientiousness. Conscientiousness is a personality trait dimension that globally describes people's propensity to be reliable, dutiful, orderly, ambitious, and cautious \citep{Jackson2010}. We (1) fit a  random forest model to predict the personality dimension of Conscientiousness, (2) compute the introduced methods for grouped feature importance (GOPFI, GPFI, GSI, LOGI, LOGO), (3) use the proposed sequential grouped feature importance procedure to investigate which groups were most important in combination, and (4) visualize the combined effect of app-usage variables with CFEPs. After the importance of individual groups has been quantified, CFEPs can be helpful to further explore the variables in these groups with regard to the criterion variable of interest (i.e., Conscientiousness) to generate new hypotheses for future research. 

The feature group \textit{app usage}, as visible in Table \ref{tab:usecase}, consistently has the highest grouped feature importance score for all calculated scores and will be explored further with the CFEPs.

 \begin{table}[ht]
\centering
\scalebox{0.7}{
\begin{tabular}{rlllll}
  \hline
  Group & GOPFI & GPFI & GSI & LOGI & LOGO \\ 
  \hline
 mobility (Mo) & -0.002 ($\pm$ 0.011) & -0.002 ($\pm$ 0.001)& 0.000 ($\pm$ 0.003)&-0.011 ($\pm$ 0.075) & 0.000 ($\pm$ 0.006) \\ 
  music (Mu) & -0.001 ($\pm$ 0.011) & 0.002 ($\pm$ 0.002)& 0.001 ($\pm$ 0.006) &-0.019 ($\pm$ 0.074)& 0.001 ($\pm$ 0.012)\\ 
  communication and social (C) & 0.000 ($\pm$ 0.008) & 0.001 ($\pm$ 0.003) & 0.004 ($\pm$ 0.006)& 0.008 ($\pm$ 0.070) & 0.001 ($\pm$ 0.010) \\ 
  overall phone usage (O) & 0.007 ($\pm$ 0.011) & 0.009 ($\pm$ 0.003) & 0.012 ($\pm$ 0.008)& 0.032 ($\pm$ 0.080) & 0.009 ($\pm$ 0.014)\\ 
  app usage (A) & 0.032 ($\pm$ 0.009) & 0.028 ($\pm$ 0.005) & 0.031 ($\pm$ 0.012)& 0.041 ($\pm$ 0.069)& 0.011 ($\pm$ 0.019) \\ 
   \hline
\end{tabular}
}
\caption{Grouped feature importance values for predicting the personality trait Conscientiousness based on MSE. GOPFI (permutation-based): Expected loss decrease, when permuting all other features except the features in a group, compared to permuting all features. GPFI (permutation-based): Expected loss increase, when all features of a group are permuted. GSI (permutation-based): Average contribution of a group to all possible combinations of groups. LOGI (refitting method): Expected loss decrease compared to guessing baseline. LOGO (refitting method): Expected loss increase compared to full model. All values were calculated by using a resampling method (10-times cross-validation).}
 \label{tab:usecase}
\end{table}

\begin{figure}
    \centering
    \includegraphics[width=0.7\textwidth]{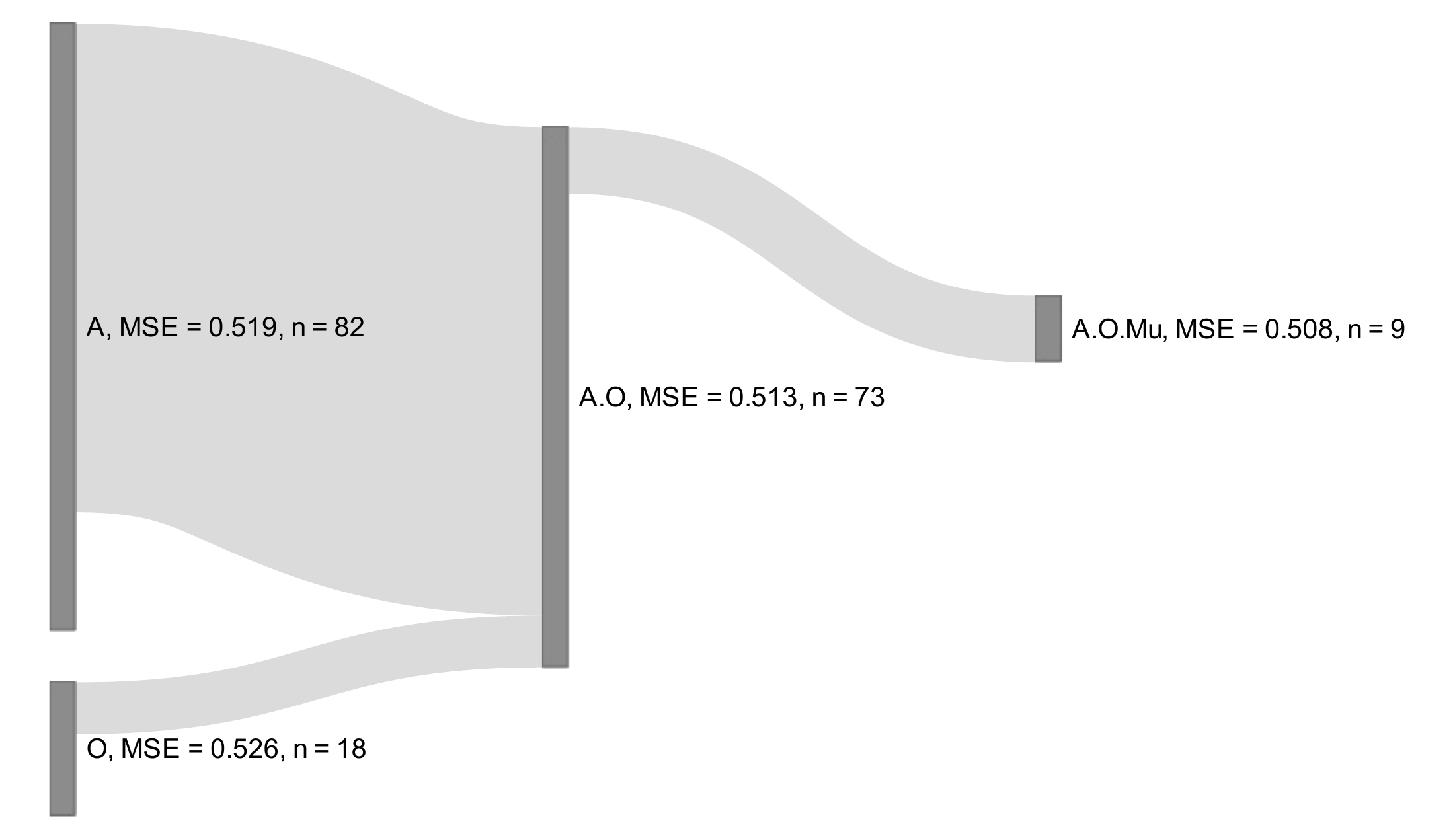}
    \caption{Sequential grouped feature importance procedure for smartphone sensor data predicting \textit{Conscientiousness}. 100 times repeated subsampling. Inner resampling strategy: 10-fold cross validation. Improvement threshold $\delta = 0.01$.  Abbreviations: app-usage (A), communication \& social (C), music (Mu), overall phone activity (O), mobility (Mo). Vertical bars show one step in the greedy forward search algorithm. Height of the vertical bars represent the number of subsampling iterations a combination of groups was chosen (for example, out of 100 subsampling iterations the group app-usage (A) was chosen 82 times as the best first group. Streams indicate proportion of iterations that additionally benefited from a consequent step. Only streams containing at least 5 iterations and better mean performance at the end are displayed.}
    \label{fig:sankeyusecase}
\end{figure}

In Figure \ref{fig:sankeyusecase}, we show a sequential procedure for our personality prediction example. The figure shows that the groups \textit{overall phone usage} and \textit{app usage} lead to the best model performance if used alone and in many cases to even better performances if combined. The figure also suggests that the initial usage of the app usage more often leads to the smallest expected loss, if only one group can be used (mean MSE = 0.519).
For a practical application, this would indicate that if you can only collect one type of feature from smartphones to predict the personality trait Conscientiousness, features on app usage should be used. If two groups of data can be collected, overall phone usage should also be added (mean MSE = 0.513). Finally, the plot indicates that in some cases (n = 9), the additional consideration of music listening behaviors in the model could lead to additional, small improvements of the expected loss (mean MSE= 0.508). Interestingly, the feature group \textit{music} alone shows very low (or even negative) grouped feature importance scores.

\begin{figure}
    \centering
    \includegraphics[width=0.7\textwidth]{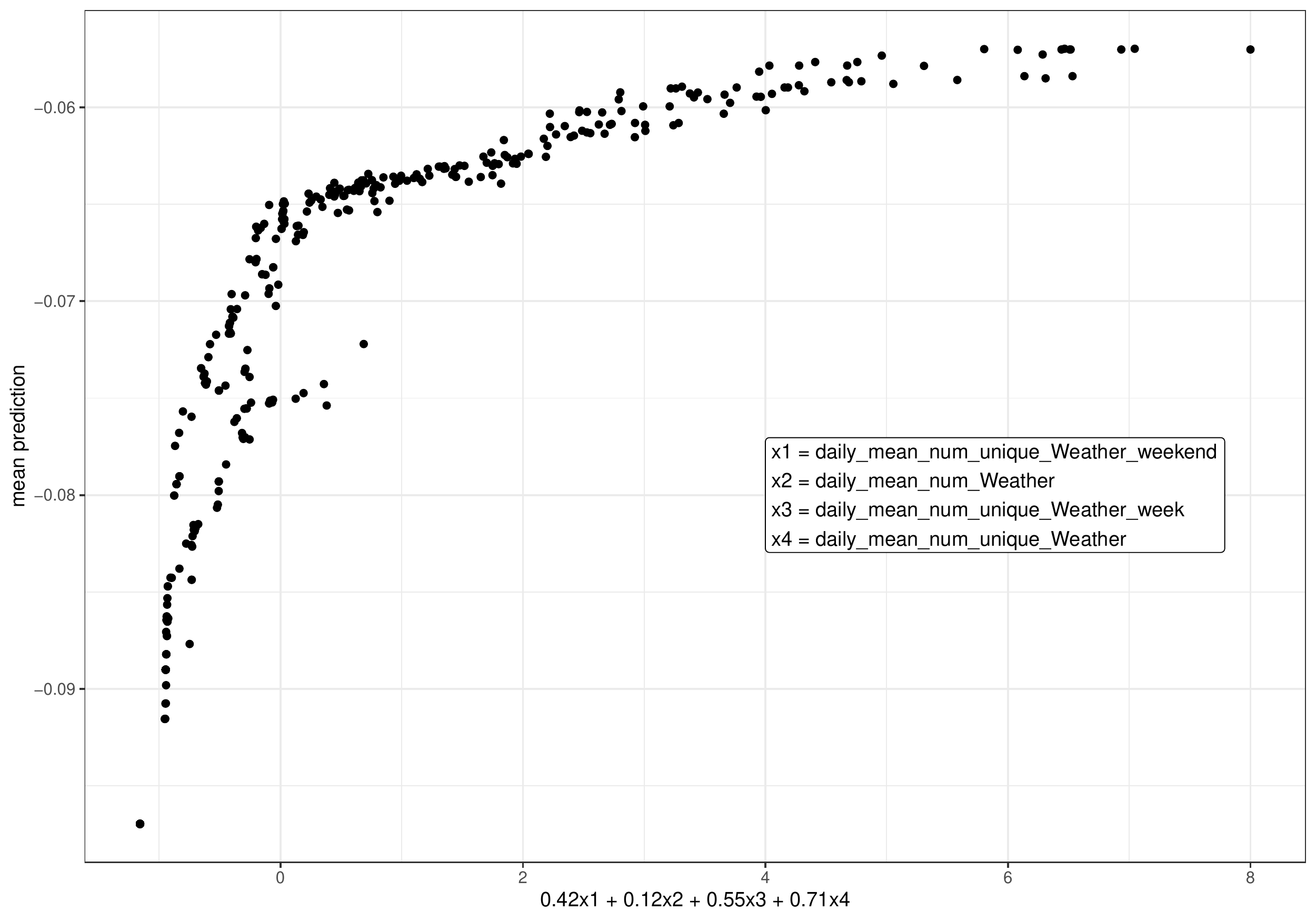}
    \caption{CFEP for the prediction of the personality trait Conscientiousness. CFEP was calculated for the group \textit{app usage} and the first principal component was chosen.}
    \label{fig:lpdpusecase1}
\end{figure}

To additionally explore meaningful and predictive directions in the feature-space of the app usage group, we use a CFEP for visualization. Figure \ref{fig:lpdpusecase1} shows that combinations of higher values in features on Weather app usage on average lead to higher mean values in the personality trait Conscientiousness. The increased frequency in weather app usage could signify the preparedness of conscientious people for future eventualities  \citep[e.g., bad weather;][]{Jackson2010}.

\FloatBarrier


\section{Conclusion}
\label{sec:conclusion}


We introduced various techniques to analyze the importance and effect of user-defined feature groups on predictions of ML models. We provided formal definitions and distinction criteria for grouped feature importance methods and distinguished between permutation- and refitting-based methods. 
For both approaches, we defined 
two calculation strategies that either start with a null model or with the full model. 
Based on these two definitions, we introduced Shapley importance scores for groups which we defined for permutation as well as refitting methods.
Moreover, we introduced a sequential grouped feature importance procedure to find good and stable combinations of feature groups. 
To contrast the newly proposed methods with existing ones, we compared them for different scenarios. The key recommendations for the user can be summarized for four scenarios: (1) If high correlations between groups are present, refitting methods should be preferred over permutation methods since they often deliver more meaningful results in these scenarios. Moreover, if the number of groups is reasonably small, refitting methods become computationally feasible. (2) If a sparse set of feature groups is of interest (e.g. due to data availability), the introduced sequential procedure can be useful. It provides insights regarding the most important groups, which sparse group combinations are stable in the sense that they are frequently selected and achieve a good performance. These criteria can be critically informative in situations where feature groups have to be obtained from different data sources that are associated with further costs. (3) If the correlation strengths of features within groups are very diverse, all of the introduced methods might fail to reflect the true underlying importance of the feature groups. The size of this effect depends heavily on how well the model captures the true underlying relationship between features. Especially when using random forests, we showed that all of the methods lead to misleading results. (4) Groups with many features might tend to have a higher grouped importance score than groups with fewer features. Normalizing the grouped importance score leads to an average score per feature. However, this might result in choosing groups with grouped scores being smaller than those of other groups and hence choosing groups that contain less information than others. When using GSI, users can extract additional feature-level information to gain more insights into the group scores. Specifically, we showed that single feature Shapley importance scores add up to GSI when no higher-order interactions between groups are present. 

We also proposed the CFEP, which is another global interpretation method that allows to visualize the combined effect of multiple features on the prediction of an ML model. By applying a supervised SPCA, we received more meaningful and interpretable results for the final CFEP than for its unsupervised counterpart. Although we only considered numeric feature spaces in all our scenarios and the real data example, all our methods are in general also applicable to mixed feature spaces. However, in presence of categorical features a suitable dimension reduction method for CFEP has to be chosen.


Here, we have focused on knowledge-driven feature groupings. However, the introduced methods could also be applied to data-driven groups (e.g., via shared variance). Obviously, their interpretation is only meaningful if groups can be described by some underlying factor. This might be a good application for interpretable latent variables to find causal relationships between feature groups and predictions of ML models. Also with regard to highly correlated feature groups that cannot be grouped naturally, a data-driven approach might be more suitable. 

We hope that this article provides a helpful reference for researchers in selecting appropriate interpretation methods when features can be grouped and that it inspires future research in this area. 



\begin{acknowledgements}
This work has been partially supported by the German Federal Ministry of Education and Research (BMBF) under Grant No. 01IS18036A, by the Bavarian State Ministry of Science and the Arts in the framework of the Centre Digitisation.Bavaria (ZD.B), a Google research grant, the LMU-excellence initiative, and the National Science Foundation (NSF) Award SES-1758835. The authors of this work take full responsibilities for its content.
\end{acknowledgements}

%
%

\bibliographystyle{spbasic}      

\newpage

\vskip 0.2in
\bibliography{Bib}

\begin{thebibliography}{75}
\providecommand{\natexlab}[1]{#1}
\providecommand{\url}[1]{{#1}}
\providecommand{\urlprefix}{URL }
\expandafter\ifx\csname urlstyle\endcsname\relax
  \providecommand{\doi}[1]{DOI~\discretionary{}{}{}#1}\else
  \providecommand{\doi}{DOI~\discretionary{}{}{}\begingroup
  \urlstyle{rm}\Url}\fi
\providecommand{\eprint}[2][]{\url{#2}}

\bibitem[{Allaire et~al.(2017)Allaire, Gandrud, Russell, and
  Yetman}]{networkD3}
Allaire J, Gandrud C, Russell K, Yetman C (2017) {networkD3}: {D3} {JavaScript}
  network graphs from {R}.
  \urlprefix\url{https://CRAN.R-project.org/package=networkD3}, {R} package
  version 0.4

\bibitem[{Amoukou et~al.(2021)Amoukou, Brunel, and
  Salaün}]{amoukou2021shapley}
Amoukou SI, Brunel NJB, Salaün T (2021) The shapley value of coalition of
  variables provides better explanations. \eprint{arXiv:2103.13342}

\bibitem[{Apley and Zhu(2019)}]{apley2016visualizing}
Apley DW, Zhu J (2019) Visualizing the effects of predictor variables in black
  box supervised learning models. \eprint{arXiv:1612.08468}

\bibitem[{Bair et~al.(2006)Bair, Hastie, Paul, and Tibshirani}]{bair2006}
Bair E, Hastie T, Paul D, Tibshirani R (2006) Prediction by supervised
  principal components. Journal of the American Statistical Association
  101(473):119--137

\bibitem[{Barshan et~al.(2011)Barshan, Ghodsi, Azimifar, and
  Jahromi}]{barshan2011supervised}
Barshan E, Ghodsi A, Azimifar Z, Jahromi MZ (2011) Supervised principal
  component analysis: Visualization, classification and regression on subspaces
  and submanifolds. Pattern Recognition 44(7):1357--1371

\bibitem[{Berk et~al.(2009)Berk, Sherman, Barnes, Kurtz, and
  Ahlman}]{berk2009forecasting}
Berk R, Sherman L, Barnes G, Kurtz E, Ahlman L (2009) Forecasting murder within
  a population of probationers and parolees: A high stakes application of
  statistical learning. Journal of the Royal Statistical Society: Series A
  (Statistics in Society) 172(1):191--211

\bibitem[{Breiman(2001)}]{Breiman2001}
Breiman L (2001) Random forests. Machine Learning 45(1):5--32

\bibitem[{Brenning(2021)}]{brenning2021transforming}
Brenning A (2021) Transforming feature space to interpret machine learning
  models. arXiv:210404295

\bibitem[{Caputo et~al.(2002)Caputo, Sim, Furesjö, and Smola}]{caputo:2002}
Caputo B, Sim K, Furesjö F, Smola A (2002) Appearance-based object recognition
  using svms: Which kernel should {I} use. In: Proc of NIPS workshop on
  statistical methods for computational experiments in visual processing and
  computer vision, Red Hook, NY, USA

\bibitem[{Casalicchio et~al.(2019)Casalicchio, Molnar, and Bischl}]{casa2019}
Casalicchio G, Molnar C, Bischl B (2019) Visualizing the Feature Importance for
  Black Box Models, Springer International Publishing. Machine Learning and
  Knowledge Discovery in Databases, pp 655--670

\bibitem[{Chakraborty and Pal(2008)}]{Chakraborty2008}
Chakraborty D, Pal NR (2008) Selecting useful groups of features in a
  connectionist framework. IEEE Transactions on Neural Networks 19(3):381--396

\bibitem[{Cohen et~al.(2005)Cohen, Ruppin, and Dror}]{Shay:2005}
Cohen SB, Ruppin E, Dror G (2005) Feature selection based on the shapley value.
  In: Kaelbling LP, Saffiotti A (eds) IJCAI-05, Proceedings of the Nineteenth
  International Joint Conference on Artificial Intelligence, Edinburgh,
  Scotland, UK, July 30 - August 5, 2005, Professional Book Center, pp 665--670

\bibitem[{Covert et~al.(2020)Covert, Lundberg, and Lee}]{covert2020SAGE}
Covert I, Lundberg SM, Lee SI (2020) Understanding global feature contributions
  with additive importance measures. Advances in Neural Information Processing
  Systems 33

\bibitem[{Eckart and Young(1936)}]{eckart1936approximation}
Eckart C, Young G (1936) The approximation of one matrix by another of lower
  rank. Psychometrika 1(3):211--218

\bibitem[{Fisher et~al.(2019)Fisher, Rudin, and Dominici}]{modelreliance2018}
Fisher A, Rudin C, Dominici F (2019) All models are wrong, but many are useful:
  Learning a variable's importance by studying an entire class of prediction
  models simultaneously. Journal of Machine Learning Research 20(177):1--81

\bibitem[{Friedman et~al.(2010)Friedman, Hastie, and
  Tibshirani}]{friedman2010note}
Friedman J, Hastie T, Tibshirani R (2010) A note on the group lasso and a
  sparse group lasso. arXiv:10010736

\bibitem[{Friedman(2001)}]{friedman2001greedy}
Friedman JH (2001) Greedy function approximation: A gradient boosting machine.
  Annals of statistics pp 1189--1232

\bibitem[{Fukumizu et~al.(2004)Fukumizu, Bach, and
  Jordan}]{fukumizu2004dimensionality}
Fukumizu K, Bach FR, Jordan MI (2004) Dimensionality reduction for supervised
  learning with reproducing kernel hilbert spaces. Journal of Machine Learning
  Research 5(Jan):73--99

\bibitem[{Goldstein et~al.(2013)Goldstein, Kapelner, Bleich, and
  Pitkin}]{icecurves}
Goldstein A, Kapelner A, Bleich J, Pitkin E (2013) Peeking inside the black
  box: Visualizing statistical learning with plots of individual conditional
  expectation. Journal of Computational and Graphical Statistics 24

\bibitem[{Gregorova et~al.(2018)Gregorova, Kalousis, and
  Marchand{-}Maillet}]{gregorova:2018}
Gregorova M, Kalousis A, Marchand{-}Maillet S (2018) Structured nonlinear
  variable selection. In: Globerson A, Silva R (eds) Proceedings of the
  Thirty-Fourth Conference on Uncertainty in Artificial Intelligence, {UAI}
  2018, Monterey, California, USA, August 6-10, 2018, {AUAI} Press, pp 23--32

\bibitem[{Gregorutti et~al.(2015)Gregorutti, Michel, and
  Saint-Pierre}]{Gregorutti2015}
Gregorutti B, Michel B, Saint-Pierre P (2015) Grouped variable importance with
  random forests and application to multiple functional data analysis.
  Computational Statistics {\&} Data Analysis 90:15--35

\bibitem[{Gretton et~al.(2005)Gretton, Bousquet, Smola, and
  Sch{\"o}lkopf}]{gretton2005measuring}
Gretton A, Bousquet O, Smola A, Sch{\"o}lkopf B (2005) Measuring statistical
  dependence with hilbert-schmidt norms. In: International conference on
  algorithmic learning theory, Springer, pp 63--77

\bibitem[{Guyon et~al.(2002)Guyon, Weston, Barnhill, and
  Vapnik}]{guyon2002gene}
Guyon I, Weston J, Barnhill S, Vapnik V (2002) Gene selection for cancer
  classification using support vector machines. Machine learning
  46(1-3):389--422

\bibitem[{Harari et~al.(2015)Harari, Gosling, Wang, and Campbell}]{Harari2015}
Harari GM, Gosling SD, Wang R, Campbell AT (2015) Capturing situational
  information with smartphones and mobile sensing methods. European Journal of
  Personality 29(5):509--511

\bibitem[{Harari et~al.(2016)Harari, Lane, Wang, Crosier, Campbell, and
  Gosling}]{Harari2016}
Harari GM, Lane ND, Wang R, Crosier BS, Campbell AT, Gosling SD (2016) Using
  smartphones to collect behavioral data in psychological science:
  Opportunities, practical considerations, and challenges. Perspectives on
  Psychological Science 11(6):838--854

\bibitem[{Harari et~al.(2017)Harari, M{\"{u}}ller, Aung, and
  Rentfrow}]{Harari2017}
Harari GM, M{\"{u}}ller SR, Aung MS, Rentfrow PJ (2017) Smartphone sensing
  methods for studying behavior in everyday life. Current Opinion in Behavioral
  Sciences 18:83--90

\bibitem[{Harari et~al.(2019)Harari, M{\"{u}}ller, Stachl, Wang, Wang,
  B{\"{u}}hner, Rentfrow, Campbell, and Gosling}]{Harari2019}
Harari GM, M{\"{u}}ller SR, Stachl C, Wang R, Wang W, B{\"{u}}hner M, Rentfrow
  PJ, Campbell AT, Gosling SD (2019) Sensing sociability: Individual
  differences in young adults' conversation, calling, texting, and app use
  behaviors in daily life. Journal of Personality and Social Psychology

\bibitem[{He and Yu(2010)}]{He2010}
He Z, Yu W (2010) Stable Feature Selection for Biomarker Discovery, vol~34,
  Computational Biology and Chemistry, pp 215--225

\bibitem[{Hein and Bousquet(2004)}]{hein2004kernels}
Hein M, Bousquet O (2004) Kernels, Associated Structures and Generalizations,
  Max Planck Institute for Biological Cybernetics

\bibitem[{Hooker(2007)}]{hooker2007generalized}
Hooker G (2007) Generalized functional anova diagnostics for high-dimensional
  functions of dependent variables. Journal of Computational and Graphical
  Statistics 16(3):709--732

\bibitem[{Hooker and Mentch(2019)}]{hooker2019}
Hooker G, Mentch L (2019) Please stop permuting features: An explanation and
  alternatives. \eprint{arXiv:1905.03151}

\bibitem[{Jackson et~al.(2010)Jackson, Wood, Bogg, Walton, Harms, and
  Roberts}]{Jackson2010}
Jackson JJ, Wood D, Bogg T, Walton KE, Harms PD, Roberts BW (2010) What do
  conscientious people do? development and validation of the behavioral
  indicators of conscientiousness (bic). Journal of Research in Personality
  44(4):501--511

\bibitem[{Lei et~al.(2018)Lei, G’Sell, Rinaldo, Tibshirani, and
  Wasserman}]{Jing2018LOCO}
Lei J, G’Sell M, Rinaldo A, Tibshirani RJ, Wasserman L (2018)
  Distribution-free predictive inference for regression. Journal of the
  American Statistical Association 113(523):1094--1111

\bibitem[{Lipton(2018)}]{lipton2018mythos}
Lipton ZC (2018) The mythos of model interpretability: In machine learning, the
  concept of interpretability is both important and slippery. Queue
  16(3):31--57

\bibitem[{Lundberg and Lee(2017)}]{lundberg:2017}
Lundberg SM, Lee SI (2017) A unified approach to interpreting model
  predictions. In: Proceedings of the 31st International Conference on Neural
  Information Processing Systems, Curran Associates Inc., Red Hook, NY, USA,
  NIPS'17, p 4768–4777

\bibitem[{Lundberg et~al.(2018)Lundberg, Erion, and Lee}]{lundberg:2018}
Lundberg SM, Erion GG, Lee S (2018) Consistent individualized feature
  attribution for tree ensembles. CoRR abs/1802.03888

\bibitem[{Meier et~al.(2008)Meier, Van De~Geer, and
  B{\"u}hlmann}]{meier2008group}
Meier L, Van De~Geer S, B{\"u}hlmann P (2008) The group lasso for logistic
  regression. Journal of the Royal Statistical Society: Series B (Statistical
  Methodology) 70(1):53--71

\bibitem[{Meinshausen and B{\"u}hlmann(2010)}]{meinshausen2010stability}
Meinshausen N, B{\"u}hlmann P (2010) Stability selection. Journal of the Royal
  Statistical Society: Series B (Statistical Methodology) 72(4):417--473

\bibitem[{de~Mijolla et~al.(2020)de~Mijolla, Frye, Kunesch, Mansir, and
  Feige}]{mijolla:2020}
de~Mijolla D, Frye C, Kunesch M, Mansir J, Feige I (2020) Human-interpretable
  model explainability on high-dimensional data. CoRR abs/2010.07384

\bibitem[{Miller(2012)}]{Miller2012}
Miller G (2012) The smartphone psychology manifesto. Perspectives on
  Psychological Science 7(3):221--237

\bibitem[{Molnar(2019)}]{molnar2019IML}
Molnar C (2019) Interpretable Machine Learning.
  \url{https://christophm.github.io/interpretable-ml-book/}

\bibitem[{Molnar et~al.(2020)Molnar, König, Bischl, and
  Casalicchio}]{molnar2020}
Molnar C, König G, Bischl B, Casalicchio G (2020) Model-agnostic feature
  importance and effects with dependent features -- a conditional subgroup
  approach. \eprint{arXiv:2006.04628}

\bibitem[{Nicodemus et~al.(2010)Nicodemus, Malley, Strobl, and
  Ziegler}]{nicodemus2010}
Nicodemus K, Malley J, Strobl C, Ziegler A (2010) The behaviour of random
  forest permutation-based variable importance measures under predictor
  correlation. BMC Bioinformatics pp 11--110

\bibitem[{Onnela and Rauch(2016)}]{Onnela2016}
Onnela JP, Rauch SL (2016) Harnessing smartphone-based digital phenotyping to
  enhance behavioral and mental health. Neuropsychopharmacology
  41(7):1691--1696

\bibitem[{Park et~al.(2006)Park, Hastie, and Tibshirani}]{Park2006}
Park MY, Hastie T, Tibshirani R (2006) Averaged gene expressions for
  regression. Biostatistics 8(2):212--227

\bibitem[{Pfister et~al.(2017)Pfister, B\"{u}hlmann, Sch\"{o}lkopf, and
  Peters}]{Pfister2017}
Pfister N, B\"{u}hlmann P, Sch\"{o}lkopf B, Peters J (2017) Kernel-based tests
  for joint independence. Journal of the Royal Statistical Society: Series B
  (Statistical Methodology) 80(1):5--31

\bibitem[{Rachuri et~al.(2010)Rachuri, Musolesi, Mascolo, Rentfrow, Longworth,
  and Aucinas}]{Rachuri2010}
Rachuri KK, Musolesi M, Mascolo C, Rentfrow PJ, Longworth C, Aucinas A (2010)
  Emotionsense: A mobile phones based adaptive platform for experimental social
  psychology research. In: UbiComp'10 - Proceedings of the 2010 ACM Conference
  on Ubiquitous Computing

\bibitem[{Raento et~al.(2009)Raento, Oulasvirta, and Eagle}]{Raento2009}
Raento M, Oulasvirta A, Eagle N (2009) Smartphones: An emerging tool for social
  scientists. Sociological Methods {\&} Research 37(3):426--454

\bibitem[{Saeb et~al.(2016)Saeb, Lattie, Schueller, Kording, and
  Mohr}]{Saeb2016}
Saeb S, Lattie EG, Schueller SM, Kording KP, Mohr DC (2016) The relationship
  between mobile phone location sensor data and depressive symptom severity.
  PeerJ 4:e2537

\bibitem[{Schoedel et~al.(2018)Schoedel, Au, V{\"{o}}lkel, Lehmann, Becker,
  B{\"{u}}hner, Bischl, Hussmann, and Stachl}]{Schoedel2018}
Schoedel R, Au Q, V{\"{o}}lkel ST, Lehmann F, Becker D, B{\"{u}}hner M, Bischl
  B, Hussmann H, Stachl C (2018) Digital footprints of sensation seeking.
  Zeitschrift f{\"{u}}r Psychologie 226(4):232--245

\bibitem[{Schoedel et~al.(2020)Schoedel, Pargent, Au, V{\"{o}}lkel, Schuwerk,
  B{\"{u}}hner, and Stachl}]{Schoedel2020}
Schoedel R, Pargent F, Au Q, V{\"{o}}lkel ST, Schuwerk T, B{\"{u}}hner M,
  Stachl C (2020) To challenge the morning lark and the night owl: Using
  smartphone sensing data to investigate day–night behaviour patterns.
  European Journal of Personality p per.2258

\bibitem[{Schuwerk et~al.(2019)Schuwerk, Kaltefleiter, Au, Hoesl, and
  Stachl}]{Schuwerk2019}
Schuwerk T, Kaltefleiter LJ, Au JQ, Hoesl A, Stachl C (2019) Enter the wild:
  Autistic traits and their relationship to mentalizing and social interaction
  in everyday life. Journal of Autism and Developmental Disorders pp 1--16

\bibitem[{Seedorff and Brown(2021)}]{seedorff2021totalvis}
Seedorff N, Brown G (2021) totalvis: A principal components approach to
  visualizing total effects in black box models. SN Computer Science 2(3):1--12

\bibitem[{Servia-Rodr{\'{i}}guez et~al.(2017)Servia-Rodr{\'{i}}guez, Rachuri,
  Mascolo, Rentfrow, Lathia, and Sandstrom}]{ServiaRodriguez2017}
Servia-Rodr{\'{i}}guez S, Rachuri KK, Mascolo C, Rentfrow PJ, Lathia N,
  Sandstrom GM (2017) Mobile sensing at the service of mental well-being: A
  large-scale longitudinal study. In: 26th International World Wide Web
  Conference, WWW 2017, International World Wide Web Conferences Steering
  Committee, pp 103--112

\bibitem[{Shapley(1953)}]{shapley1953value}
Shapley LS (1953) A value for n-person games. Contributions to the Theory of
  Games 2(28):307--317

\bibitem[{Sharifzadeh et~al.(2017)Sharifzadeh, Ghodsi, Clemmensen, and
  Ersb{\o}ll}]{sharifzadeh2017sparse}
Sharifzadeh S, Ghodsi A, Clemmensen LH, Ersb{\o}ll BK (2017) Sparse supervised
  principal component analysis (sspca) for dimension reduction and variable
  selection. Engineering Applications of Artificial Intelligence 65:168--177

\bibitem[{Shipp et~al.(2002)Shipp, Ross, Tamayo, Weng, Kutok, Aguiar,
  Gaasenbeek, Angelo, Reich, Pinkus et~al.}]{shipp2002diffuse}
Shipp MA, Ross KN, Tamayo P, Weng AP, Kutok JL, Aguiar RC, Gaasenbeek M, Angelo
  M, Reich M, Pinkus GS, et~al. (2002) Diffuse large b-cell lymphoma outcome
  prediction by gene-expression profiling and supervised machine learning.
  Nature medicine 8(1):68--74

\bibitem[{Song et~al.(2007)Song, Smola, Gretton, Borgwardt, and
  Bedo}]{song2007supervised}
Song L, Smola A, Gretton A, Borgwardt KM, Bedo J (2007) Supervised feature
  selection via dependence estimation. In: Proceedings of the 24th
  international conference on Machine learning, pp 823--830

\bibitem[{Song et~al.(2012)Song, Smola, Gretton, Bedo, and
  Borgwardt}]{song2012feature}
Song L, Smola A, Gretton A, Bedo J, Borgwardt K (2012) Feature selection via
  dependence maximization. Journal of Machine Learning Research
  13(May):1393--1434

\bibitem[{Stachl et~al.(2017)Stachl, Hilbert, Au, Buschek, {De Luca}, Bischl,
  Hussmann, and B{\"{u}}hner}]{Stachl2017}
Stachl C, Hilbert S, Au JQ, Buschek D, {De Luca} A, Bischl B, Hussmann H,
  B{\"{u}}hner M (2017) Personality traits predict smartphone usage. European
  Journal of Personality 31(6):701--722

\bibitem[{Stachl et~al.(2020{\natexlab{a}})Stachl, Au, Schoedel, Gosling,
  Harari, Buschek, Theres, Völkel, Schuwerk, Oldemeier, Ullmann, Hussmann,
  Bischl, and Bühner}]{Stachl2020}
Stachl C, Au Q, Schoedel R, Gosling SD, Harari GM, Buschek D, Theres S,
  Völkel, Schuwerk T, Oldemeier M, Ullmann T, Hussmann H, Bischl B, Bühner M
  (2020{\natexlab{a}}) Predicting personality from patterns of behavior
  collected with smartphones. Proceedings of the National Academy of Sciences

\bibitem[{Stachl et~al.(2020{\natexlab{b}})Stachl, Pargent, Hilbert, Harari,
  Schoedel, Vaid, Gosling, and B{\"{u}}hner}]{Stachl2020ML}
Stachl C, Pargent F, Hilbert S, Harari GM, Schoedel R, Vaid S, Gosling SD,
  B{\"{u}}hner M (2020{\natexlab{b}}) Personality research and assessment in
  the era of machine learning. European Journal of Personality p per.2257

\bibitem[{Strobl et~al.(2008)Strobl, Boulesteix, Kneib, Augustin, and
  Zeileis}]{Strobl2008}
Strobl C, Boulesteix AL, Kneib T, Augustin T, Zeileis A (2008) Conditional
  variable importance for random forests. BMC bioinformatics 9:307

\bibitem[{Thom{\'{e}}e(2018)}]{Thomee2018}
Thom{\'{e}}e S (2018) Mobile phone use and mental health. a review of the
  research that takes a psychological perspective on exposure. International
  Journal of Environmental Research and Public Health 15(12):2692

\bibitem[{Tibshirani(1996)}]{tibshirani1996regression}
Tibshirani R (1996) Regression shrinkage and selection via the lasso. Journal
  of the Royal Statistical Society: Series B (Methodological) 58(1):267--288

\bibitem[{Toloşi and Lengauer(2011)}]{Tolosi2011}
Toloşi L, Lengauer T (2011) Classification with correlated features:
  Unreliability of feature ranking and solutions. Bioinformatics
  27(14):1986--1994

\bibitem[{Tripathi et~al.(2020)Tripathi, Hemachandra, and
  Trivedi}]{tripathi2020interpretable}
Tripathi S, Hemachandra N, Trivedi P (2020) Interpretable feature subset
  selection: A shapley value based approach. In: Proceedings of 2020 IEEE
  International Conference on Big Data, Special Session on Explainable
  Artificial Intelligence in Safety Critical Systems

\bibitem[{Valentin et~al.(2020)Valentin, Harkotte, and
  Popov}]{valentin2020interpreting}
Valentin S, Harkotte M, Popov T (2020) Interpreting neural decoding models
  using grouped model reliance. PLOS Computational Biology 16(1):e1007148

\bibitem[{Watson and Wright(2019)}]{Watson2019}
Watson DS, Wright MN (2019) Testing {Conditional} {Independence} in
  {Supervised} {Learning} {Algorithms}. arXiv:190109917

\bibitem[{Williamson et~al.(2020)Williamson, Gilbert, Simon, and
  Carone}]{williamson2020unified}
Williamson BD, Gilbert PB, Simon NR, Carone M (2020) A unified approach for
  inference on algorithm-agnostic variable importance. arXiv:200403683

\bibitem[{Witten and Tibshirani(2020)}]{RPMA}
Witten D, Tibshirani R (2020) PMA: Penalized Multivariate Analysis. {R} package
  version 1.2.1

\bibitem[{Witten et~al.(2009)Witten, Tibshirani, and
  Hastie}]{witten2009penalized}
Witten DM, Tibshirani R, Hastie T (2009) A penalized matrix decomposition, with
  applications to sparse principal components and canonical correlation
  analysis. Biostatistics 10(3):515--534

\bibitem[{Wold et~al.(1984)Wold, Albano, Dunn, Edlund, Esbensen, Geladi,
  Hellberg, Johansson, Lindberg, and Sj{\"o}str{\"o}m}]{Wold1984}
Wold S, Albano C, Dunn WJ, Edlund U, Esbensen K, Geladi P, Hellberg S,
  Johansson E, Lindberg W, Sj{\"o}str{\"o}m M (1984) Multivariate Data Analysis
  in Chemistry, Springer Netherlands, Dordrecht, pp 17--95

\bibitem[{Yarkoni and Westfall(2017)}]{Yarkoni2017}
Yarkoni T, Westfall J (2017) Choosing prediction over explanation in
  psychology: Lessons from machine learning. Perspectives on Psychological
  Science 12(6):1100--1122

\bibitem[{Yuan and Lin(2006)}]{Yuan2006}
Yuan M, Lin Y (2006) Model selection and estimation in regression with grouped
  variables. Journal of the Royal Statistical Society: Series B (Statistical
  Methodology) 68(1):49--67

\end{thebibliography}

\newpage
\appendix

\section{Shapley Importance}

\label{sec:shapleyproof}

Assume, that the value function for a coalition $S\subset\{x_1, ..., x_p\}$ can be broken down into main and interaction effects:
\begin{equation*}
v(S) = \sum_{x_i \in S} v(x_i) + \sum_{i_1 \neq i_2}\epsilon_{i_1i_2} + \sum_{i_1\neq i_2 \neq i_3} \epsilon_{i_1i_2i_3} + ... ,   
\end{equation*}
the Shapley importance of a single feature $x_1$ can be written as
\begin{equation*}
    \phi(x_1) = v(x_1) + \frac{1}{2} \left(\sum_{i \neq 1}^p \epsilon_{1i}\right)+ \frac{1}{3}\left(\sum_{i \neq j \neq 1}^p \epsilon_{1ij} \right) + ... + \frac{1}{p}\epsilon_{1...p}.
\end{equation*}

\textbf{Proof:}

Let $N = \{x_2, ..., x_p\}$. The general formula for the Shapley importance is given by:
\begin{equation}
    \label{eq:shapleygeneral}
    \phi_p(x_1) = \sum_{S \subset N\backslash\{x_1\}}\frac{(p - 1 - |S|)! \cdot |S|!}{p!}\left( v(S \cup \{x_1\} ) - v(S)\right)
\end{equation}
With assumption (\ref{eq:interaction}) the term $v(S \cup \{x_1\} ) - v(S)$ will reduce to:
\begin{equation}
    \label{eq:proofshapley1}
    v(S \cup \{x_1\} ) - v(S) = v(x_1) + \sum_{i_1 \neq 1}^p \epsilon_{1i_1} + ... + \sum_{i_1 \neq ... \neq i_{|S|}\neq 1}^p \epsilon_{1i_1...i_{|S|}}
\end{equation}
It is the sum of $v(x_1)$ and all interactions with feature $x_1$ of sizes $2,...,|S| + 1$. All other terms without feature $x_1$ cancel out.

Equation (\ref{eq:shapleygeneral}) consists of many summands of the form (\ref{eq:proofshapley1}).
The term  $v(x_1)$ appears once for every subset $S \subset N\backslash\{x_1\}$. There are $\binom{p-1}{|S|}$ different subsets of size $|S|$.
Only looking at the summands with the term $v(x_1)$, equation (\ref{eq:shapleygeneral}) reduces to
\begin{align}
\sum_{|S| = 0}^{p-1} \frac{(p-1-|S|)!\cdot |S|!}{p!} \binom{p-1}{|S|} v(x_1) =  v(x_1).
\end{align}
For the interaction terms, we first start counting the interaction term $\epsilon_{12}$ of size $2$, as an example.
For $|S| = 0$, there are zero terms of $\epsilon_{12}$. For $|S| = 1$, the term $\epsilon_{12}$ only appears once, when $S = \{ x_2 \}$. For $|S| = 2$, the term $\epsilon_{12}$ appears $p-2$ times, once for each subset $S = \{x_2, x_j\}$, for $3 \leq j \leq p$. For $|S| = 3$, we have $\binom{p-2}{2}$ times the term $\epsilon_{12}$, again, once for each subset $S = \{x_2, x_j, x_k\}$, for $3 \leq j \neq k \leq p$. This pattern goes on until there are $\binom{p-2}{p-2}$ terms of $\epsilon_{12}$ for $|S| = p-1$.
Now, we look at the interaction terms $\epsilon_{1i_1...i_{k-1}}$ of size $k$. Following the pattern, which we just derived, there are zero terms of  $\epsilon_{1i_1...i_{k-1}}$ for $|S| \leq k-2$ and $\binom{p-k}{|S| - k + 1}$ terms of $\epsilon_{1i_1...i_{k-1}}$ for $k \leq |S| \leq p-1$. 
If we only look at the interaction terms  $\epsilon_{1i_1...i_{k-1}}$ of size $k$ and following the equation (\ref{eq:shapleygeneral}) , we get
\begin{align*}
\sum_{|S| = k-1}^{p-1} \frac{(p-1-|S|)!\cdot |S|!}{p!} \binom{p - k}{|S| - k + 1} \epsilon_{1i_1...i_{k-1}} = \frac{1}{k} \epsilon_{1i_1...i_{k-1}},
\end{align*}
which was left to show the assertion.  \hfill $\square$

\section{More Details on Dimension Reduction Techniques}
\label{sec:sspca_details}
\subsection{Principal Component Analysis}
PCA only considers the data matrix $\mathbf X$ and does not take the target vector $\mathbf Y$ into account. This procedure is thus unsupervised.

Given a centering Matrix 
\begin{equation}
    \label{centeringmatrix}
    \mathbf H = \mathbf I - n^{-1}ee^T,
\end{equation}
where $e$ is an $n$-dimensional vector of all ones. The centered matrix is $\mathbf X_C = \mathbf H\mathbf X$. The sample covariance matrix of $\mathbf X$ can be written as:
\begin{equation}
\mathbf S_{\mathbf X} := \frac{1}{n} \mathbf X_C^\intercal \mathbf X_C = \frac{1}{n} \mathbf X^\intercal \mathbf H \mathbf H \mathbf X
\label{pca}
\end{equation}
The goal is to maximize the total variance of projected data, which is equivalent to maximizing trace of the sample covariance matrix.  Eq.~\eqref{pca} can also be written as $\mathbf S_{\mathbf X} = \frac{1}{n}\sum_{i=1}^n \mathbf x_C^{(i)} \mathbf x_C^{(i)\intercal}$, where $\mathbf x_C^{(i)}$ corresponds to the $i-$th row of $\mathbf X_C$. 
By projecting each data point by some unknown vectors $\mathbf v_j, j = 1,...,p$, we get the projected variance for each $j = 1,...,p$, which is:
$$
\frac{1}{n}\sum_{i = 1}^n \mathbf v_j^\intercal \mathbf x_C^{(i)} \mathbf x_C^{(i)\intercal} \mathbf v_j = \mathbf v_j^\intercal \left(\frac{1}{n}\sum_{i = 1}^n  \mathbf x_C^{(i)} \mathbf x_C^{(i)\intercal}\right) \mathbf v_j = \mathbf v_j^\intercal \mathbf S_{\mathbf X} \mathbf v_j.
$$
Let $ \mathbf V \in \mathbb R^{p \times p}$ be the full projection matrix. The projected total variance is $tr( \mathbf V^\intercal \mathbf S_{\mathbf X}  \mathbf V)$, and by ignoring constant terms, PCA finds a solution to the problem 
\begin{equation}
\label{PCAargmax}
\underset{ \mathbf V}{\text{argmax }} tr( \mathbf V^\intercal \mathbf S_{\mathbf X}  \mathbf V) = \underset{ \mathbf V}{\text{argmax }} tr( \mathbf V^\intercal \mathbf X^\intercal \mathbf H \mathbf H \mathbf X  \mathbf V)
\end{equation}
with an Eigen decomposition of the covariance matrix $\mathbf S_{\mathbf X}$. The resulting Eigen vectors thus maximize the variation of projected data.

\subsection{Measuring Statistical Dependence with Hilbert Schmidt Norms}
In \cite{gretton2005measuring} a more generalized measure of dependence between variables X and Y was introduced:

Two random variables $X$ and $Y$ are independent if and only if any bounded continuous function of them are uncorrelated. 

In more detail, this means that any pairs $(X, Y), (X, Y^2), (X^2, Y), (cos(X), log(Y)), ...$ have to be uncorrelated. The resulting independence measure is called the Hilbert-Schmidt Independence Criterion (HSIC).
For the analysis of this independence measure, it is necessary to analyze functions on random variables. Therefore theory of Hilbert spaces and concepts of functional analysis are necessary for a thorough analysis, but they are not part of this paper.
For an extensive discussion of Hilbert spaces, especially reproducing kernel hilbert spaces (RKHS) we refer to \cite{hein2004kernels}.

Let $\mathcal F$ be a separable RKHS containing all bounded continuous functions from $\mathcal X$ to $\mathbb R$. The associated kernel shall be denoted by $ \mathbf K \in \mathbb R^{n \times n}$, with $\mathbf K_{ij} = k(x_i, x_j)$. Concurrently, let $\mathcal G$ be a separable RKHS with bounded continuous functions from $\mathcal Y$ to $\mathbb R$ and associated kernel $ \mathbf L \in \mathbb R^{n\times n}$, with $\mathbf L_{ij} = l(y_i, y_j)$.

We are particularly interested in the cross variance between $f$ and $g$:  
\begin{equation}
    Cov(f(x),g(y)) = \mathbb E_{x, y}[f(x)g(y)] - \mathbb E_x[f(x)]\mathbb E_y[g(y)]
\end{equation}
A function, which maps one element from one hilbert space to another hilbert space is called \textit{operator}. A theorem (see e.g. \cite{fukumizu2004dimensionality}) states, that there exists a unique operator $C_{X,Y}:\mathcal G \longrightarrow \mathcal{F}$ with
\begin{equation}
    \langle f, C_{x,y}(g)\rangle_\mathcal{F} = Cov(f(x),g(y)).
\end{equation}
The Hilbert-Schmidt Independence Criterion (HSIC) is defined as the squared Hilbert-Schmidt norm of the cross-covariance operator C:
\begin{equation}
    \text{HSIC}(P_{\mathcal X, \mathcal Y}, \mathcal F, \mathcal G) = ||C_{x,y}||_{HS}^2
\end{equation}
$||C_{x,y}||_{HS}^2 = 0$ if and only if the random variables $\mathcal X$ and $\mathcal Y$ are independent.
For a detailed discussion and derivation of the HSIC independence measure, we refer to 
\cite{gretton2005measuring}. The HSIC measure was used for feature selection in \cite{song2007supervised} or for supervised principal components in \cite{barshan2011supervised}.
\subsubsection{Empirical HSIC}
For a dataset $\mathcal D = \{(\mathbf x^{(i)}, y^{(i)})\}_{i = 1}^n$ the empirical HSIC is:
\begin{equation}
    HSIC(\mathcal D, F, G) = (n-1)^{-2} tr(\mathbf K\mathbf H\mathbf L\mathbf H) = (n-1)^{-2} tr(\mathbf H\mathbf K\mathbf H\mathbf L),
\end{equation}
where $\mathbf H$ is the centering matrix from (\ref{centeringmatrix}).
A high level of dependency between two kernels yields a high HSIC value.

\subsection{Supervised Sparse Principal Components}
In the process of finding interpretable latent variables, which also incorporate dependencies to a target variable, the Sparse Supervised Principal Components (SPCA), which was introduced in \cite{sharifzadeh2017sparse}, is a suitable method for our application.

 For sparse SPCA the kernel matrix $K$ ist defined as $K = XV V^\intercal X^\intercal$ with a constraint for unit length and an $L_1$ penalty for sparsity. By ignoring constant terms, we get the optimization problem:
\begin{align}
    \underset{ \mathbf V}{\text{argmax }} tr(\mathbf H\mathbf K\mathbf H\mathbf L) &= \underset{ \mathbf V}{\text{argmax }} tr(\mathbf H\mathbf X\mathbf V \mathbf V^\intercal \mathbf X^\intercal \mathbf H\mathbf L) \\
    &=
    \underset{\mathbf  V}{\text{argmax }} tr(\mathbf V^\intercal \mathbf X^\intercal \mathbf H\mathbf L\mathbf H\mathbf X\mathbf V) \label{connectionPCA}\\
    & s.t. \hspace{2pt} \mathbf V^\intercal \mathbf V = \mathbf I, \hspace{2pt}|\mathbf V| \leq c.
\end{align}
Note, that without the sparsity constraint, (\ref{connectionPCA}) reduces to (\ref{PCAargmax}), when choosing $\mathbf L = \mathbf I$. Already explained in \cite{barshan2011supervised}, PCA is a special form of their Supervised PCA, where setting $\mathbf L = \mathbf I$ is a kernel, which only captures similarity between a point and itself. Maximizing dependency between $\mathbf K$ and the identiy matrix corresponds to retaining maximal diversity between observations.

Now, an arbitrary $\mathbf L$ can be decomposed as $\mathbf L = \Delta \Delta^\intercal$, since $\mathbf L$, as a kernel matrix, is positive definite and symmetric. Defining $\Psi:=\Delta^\intercal \mathbf H \mathbf X \in \mathbb R^{n \times p}$, the objective function (\ref{connectionPCA}) can be rewritten as: 

\begin{align}
\label{rewrittenobjective}
  \underset{ \mathbf V}{\text{argmax }} tr(\mathbf V^\intercal \Psi^\intercal \Psi \mathbf V) \hspace{2pt}  s.t. \hspace{2pt} \mathbf V^\intercal \mathbf V = \mathbf I, \hspace{2pt}|\mathbf V| \leq c.
\end{align}

Using the singular value decomposition (SVD), the matrix $\Psi$ with $\text{rank}(\Psi) = m \leq n$ can be written as a product of matrices:
\begin{align}
  \Psi = \mathbf U \Lambda \mathbf V^\intercal \hspace{2pt} s.t. \hspace{2pt} \mathbf U^\intercal \mathbf U = I_n, \mathbf V\mathbf V^\intercal = I_p, \Lambda = I(\lambda_1,..., \lambda_m,0,...,0),
\end{align}
where $\mathbf U \in \mathbb R^{n \times n}$ and $\mathbf V \in \mathbb R^{p \times p}$ are orthogonal matrices, and $\Lambda \in \mathbb R^{n \times p}$ is a diagonal matrix, with descending diagonal entries $\lambda_1 \geq \lambda_2 \geq ...\geq \lambda_m \geq 0$.
It is easy to see that the columns of $\mathbf V$ are Eigen vectors of the matrix $\Psi^\intercal \Psi$, since the following Eigen value decomposition holds:
\begin{equation}
\label{Vev}
    \Psi^\intercal\Psi = \mathbf V \Lambda \mathbf U^\intercal \mathbf U \Lambda \mathbf V^\intercal = \mathbf V (\Lambda^2) \mathbf V^\intercal.
\end{equation}

The sparse SPCA problem (\ref{rewrittenobjective}) now becomes a matrix decomposition problem of the matrix $\Psi$, when adding an $L_1$ penalty on the matrix $\mathbf V$, since the columns of $\mathbf V$, being Eigen vectors of $\Psi^\intercal \Psi$, maximize $tr(\mathbf V^\intercal \Psi^\intercal \Psi \mathbf V)$.  

With an $L_1$ penalty on $\mathbf V$, this problem is a \textit{penalized matrix decomposition} problem (PMD, \cite{witten2009penalized}).

Recalling our original problem of finding interpretable latent variables that also depend on a target variable, the rank $m$ matrix decomposition of $\Psi$ may not be desirable.
It can be shown (e.g. \cite{eckart1936approximation}) that the best low rank ($r \leq m$) approximation of $\Psi$ is calculated by the first $r$ singular values of $\Lambda$ and the first $r$ singular vectors of $\mathbf U$ and $\mathbf V$. With $\mathbf u_i$ being the $i-$th column of $ \mathbf U$ and $\mathbf v_i$ being the $i-$th column of $\mathbf V$, the best low rank approximation can thus be written as:
\begin{equation}
    \label{frobenius}
    \sum_{i = 1}^{r}\lambda_i \mathbf u_i \mathbf v_i^\intercal =\underset{\hat \Psi}{\text{argmin}} || \Psi - \hat \Psi ||^2_F, 
\end{equation}
subject to the squared Frobenius-norm ($A\in \mathbb R^{m \times n}$: $||A||^2_F = \sum_{i = 1}^{n} \sum_{j = 1}^{m} |a_{ij}|^2$).
The following equality was demonstrated in \cite{witten2009penalized}:
\begin{equation}
  \frac{1}{2} || \Psi - \mathbf U \Lambda \mathbf V^\intercal ||^2_F = \frac{1}{2}|| \Psi||^2_F - \sum_{i = 1}^{r} \mathbf{u}_i^\intercal \Psi \mathbf v_i \lambda_i + \frac{1}{2}\sum_{i = 1}^{r} \lambda_i^2.
\end{equation}

The minimization problem (\ref{frobenius}) thus becomes a maximization problem, by ignoring the constant terms. \cite{sharifzadeh2017sparse} added additional $L_2$ constraints on $\mathbf{u}_i$ and $\mathbf v_i$, an $L_1$ constaint on $v_i$ for sparsity and an orthogonality constraint for $u_i$:

\begin{equation}
\label{sspcaPMD}
    \underset{ \mathbf u_i \mathbf v_i}{\text{argmax }} \mathbf u_i^\intercal \Psi \mathbf v_i \hspace{2pt} s.t. ||\mathbf u_i ||_2 \leq 1, ||\mathbf v_i||_2 \leq 1, ||\mathbf v_i||_1 \leq c, 
    \mathbf u_i \perp \mathbf u_1, ..., \mathbf u_{i-1}
\end{equation}

The $L_2$ constraints do not force unit length to avoid non convex optimization problems. \cite{witten2009penalized} discuss how to solve many penalized matrix decomposition problems of this kind. Without the orthogonality constraint, they call this particular problem PMD$(., L_1)$. The solution to this problem is discussed in detail in \cite{sharifzadeh2017sparse}.
A software implementation is available with the R-package PMA by \cite{RPMA}, which we will use for our demonstrations. Problem (\ref{sspcaPMD}) does not yield orthogonal sparse vectors $\mathbf v_i$, \cite{witten2009penalized} state that these vectors are unlikely to be very correlated, since the vectors $\mathbf v_i$ are associated with orthogonal vectors $\mathbf u_i$, $i = 1, ..., r.$

\subsubsection{Choice of the Kernel}
For sparse SPCA the kernel $\mathbf K$ has been predefined as. The choice of the kernel $\mathbf L$, however, has a decisive impact on how the dependencies are modeled. \cite{song2012feature} discuss the kernel choice for different situations. For binary classification, one may simply choose 
\begin{equation}
    l(y_i,y_j) = y_iy_j, \text{ where }y_i,y_j\in \{\pm 1\},
\end{equation}
or a weighted version, giving different weights on positive and negative labels.
For multiclass classification a possible kernel is
\begin{equation}
    l(y_i,y_j) = c_y \delta_{y_i, y_j}, \text{ where }c_y >0.
\end{equation}
For regression one can also use a linear kernel $l(y_i, y_j) = y_i,y_j$, but then only simple linear correlations between features and the target variable can be detected. A more universal choice is the radial basis function (RBF) kernel:
\begin{equation}
    l(y_i, y_j) = exp\left(- \frac{||y_i - y_j||^2}{2\sigma^2}\right).
\end{equation}
The choice of the bandwidth $2\sigma^2$ is extremely important. For example, if $2\sigma^2 \rightarrow 0$, the matrix L becomes the identity matrix. Or if $2\sigma^2 \rightarrow \infty$, all entries of $\mathbf L$ are $1$. In both cases, all relevant information of the dependency between features and the target variable is lost. Besides the bandwidth $2\sigma$, the kernel matrix $L$ depends only on the pairwise distances $||y_i - y_j|||^2$. A reasonable, and heuristically well performing \citep{Pfister2017} choice is $2\sigma^2 = \text{median} \hspace{-2pt}\left( ||y_i - y_j||^2: i>j\right)$.
However, it might also be possible and advantageous to use other kernels that are selected to be particularly efficient in detecting certain kinds of dependencies.

\subsubsection{Choice of c}
\cite{witten2009penalized} explained how PMD can be used to impute missing data. 
The main idea is simply to exclude missing entries from the maximization problem (\ref{sspcaPMD}) and impute missing values by the low rank approximation matrix $\mathbf U\Lambda \mathbf V^\intercal$. 
This procedure can also be used for finding optimal values for $c$ by a cross validation approach. 
The test data consists of leaving out some entries of the matrix  $\Psi$ (not entire rows or columns, but individual elements of the matrix), yielding a matrix with missing entries $\tilde \Psi$. For candidate values $c_i, i = 1,...,k$, calculate the PMD$(., L_1)$ and record the mean squared error over the missing elements of $\tilde \Psi$ and the estimate $\mathbf U\Lambda \mathbf V^\intercal$. The true values of the missing values of $\tilde \Psi$ are available in the original data $\Psi$.
The optimal value $c^*$ corresponds to the best candidate value $c_j$, which minimizes the mean squared error.

However, such a cross-validation approach for the search for $c$ is not always necessary. If the method is used as a descriptive method to better understand the underlying structure of the data, a small value of $c$ can be chosen to achieve a desired sparsity.
\end{document}